\documentclass[lettersize,journal]{IEEEtran}
\usepackage{amsmath,amsfonts}
\usepackage{algpseudocode}
\usepackage{algorithmicx}  
\usepackage{algorithm}
\usepackage{array}
\usepackage[caption=false,font=normalsize,labelfont=sf,textfont=sf]{subfig}
\usepackage{textcomp}
\usepackage{stfloats}
\usepackage{url}
\usepackage{verbatim}
\usepackage{graphicx}
\usepackage{cite}
\usepackage{amssymb}
\usepackage{enumitem}
\usepackage[utf8]{inputenc}
\usepackage{bm}
\usepackage{multirow}
\usepackage{tabularx} 
\usepackage{hyperref}
\usepackage{balance}
\usepackage[table,xcdraw]{xcolor}
\begin{document}

\title{High Efficiency Wiener Filter-based Point Cloud Quality Enhancement for MPEG G-PCC}

\author{Yuxuan Wei, Zehan Wang, Tian Guo, Hao Liu, Liquan Shen and Hui Yuan~\IEEEmembership{Senior Member,~IEEE}
\thanks{This work was supported in part by the National Natural Science Foundation of China under Grants 62222110, 62172259, and 62311530104, the Taishan Scholar Project of Shandong Province (tsqn202103001), the Natural Science Foundation of Shandong Province under Grant ZR2022ZD38, ZR2023QF111, ZR2024MF118 and the OPPO Research Fund. (\textit{Corresponding author: Hui Yuan.})}
\thanks{Yuxuan Wei, Zehan Wang, Tian Guo and Hui Yuan are with the School of Control Science and Engineering, Shandong University, Jinan 250061, China, and also with the Key Laboratory of Machine Intelligence and System Control, Ministry of Education, Jinan 250061, China (Email: \protect\href{mailto:sduwyuxuan@mail.sdu.edu.cn}{sduwyuxuan@mail.sdu.edu.cn};
\protect\href{mailto:wangzehan@mail.sdu.edu.cn}{wangzehan@mail.sdu.edu.cn};
\protect\href{mailto:guotiansdu@mail.sdu.edu.cn}{guotiansdu@mail.sdu.edu.cn};
\protect\href{mailto:huiyuan@sdu.edu.cn}{huiyuan@sdu.edu.cn}).}
\thanks{Hao Liu is with the School of Computer and Control Engineering, Yantai University, Yantai, 264005, China (E-mail: 
\protect\href{mailto:liuhaoxb@gmail.com}{liuhaoxb@gmail.com}).}
\thanks{Liquan Shen is with the Shanghai Institute for Advanced Communication and Data Science, Shanghai University, Shanghai 200072, China (E-mail: 
\protect\href{mailto:jsslq@shu.edu.cn}{jsslq@shu.edu.cn}).}}

\markboth{Journal of \LaTeX\ Class Files,~Vol.~XX, No.~X, December~2024}%
 {High Efficiency Wiener Filter for Attribute Quality Enhancement in MPEG G-PCC}

\makeatletter
\def\ps@IEEEtitlepagestyle{
  \def\@oddfoot{\mycopyrightnotice}
  \def\@evenfoot{}
}
\def\mycopyrightnotice{
  {\footnotesize
  \begin{minipage}{\textwidth}
  \centering
  Copyright~\copyright~2025 IEEE. Personal use of this material is permitted. However, permission to use this material for any other purposes \\ 
 must be obtained from the IEEE by sending an email to pubs-permissions@ieee.org.
  \end{minipage}
  }
}

\maketitle

\begin{abstract}
Point clouds, which directly record the geometry and attributes of scenes or objects by a large number of points, are widely used in various applications such as virtual reality and immersive communication. However, due to the huge data volume and unstructured geometry, efficient compression of point clouds is very crucial. The Moving Picture Expert Group is establishing a geometry-based point cloud compression (G-PCC) standard for both static and dynamic point clouds in recent years. Although lossy compression of G-PCC can achieve a very high compression ratio, the reconstruction quality is relatively low, especially at low bitrates. To mitigate this problem, we propose a high efficiency Wiener filter that can be integrated into the encoder and decoder pipeline of G-PCC to improve the reconstruction quality as well as the rate-distortion performance for dynamic point clouds. Specifically, we first propose a basic Wiener filter, and then improve it by introducing coefficients inheritance and variance-based point classification for the Luma component. Besides, to reduce the complexity of the nearest neighbor search during the application of the Wiener filter, we also propose a Morton code-based fast nearest neighbor search algorithm for efficient calculation of filter coefficients. Experimental results demonstrate that the proposed method can achieve average Bjøntegaard delta rates of -6.1\%, -7.3\%, and -8.0\% for Luma, Chroma Cb, and Chroma Cr components, respectively, under the condition of lossless-geometry-lossy-attributes configuration compared to the latest G-PCC encoding platform (i.e., geometry-based solid content test model version 7.0 release candidate 2) by consuming affordable computational complexity. 
\end{abstract}


\section{Introduction}
\IEEEPARstart{A} point cloud is a set of points that precisely describe the geometry and the corresponding attributes for the surface of an object or a scene \cite{ref1}. It can be used extensively in fields such as virtual reality \cite{ref2}, and metaverse \cite{ref3}. However, its data volume is very huge, challenging the current network bandwidth and storage capacity. Effective point cloud compression is vital to address this challenge as well as enhancing real-time processing capabilities \cite{ref4}. 

To promote the widespread application of point clouds, the Moving Picture Experts Group (MPEG) started the standardization of 3D point cloud compression in 2017 \cite{ref5}, leading to two branches: video-based point cloud compression (V-PCC) \cite{ref6} and geometry-based point cloud compression (G-PCC) \cite{ref7}. As the 3D digital human and metaverse is becoming more and more popular, currently, a sub-branch of G-PCC, named as geometry-based solid content test model (GeS-TM), is specifically designed for the compression of dynamic solid (i.e., denser) point clouds \cite{ref8}. 

\begin{figure}[!t]
\centering
\includegraphics[width=0.44\textwidth]{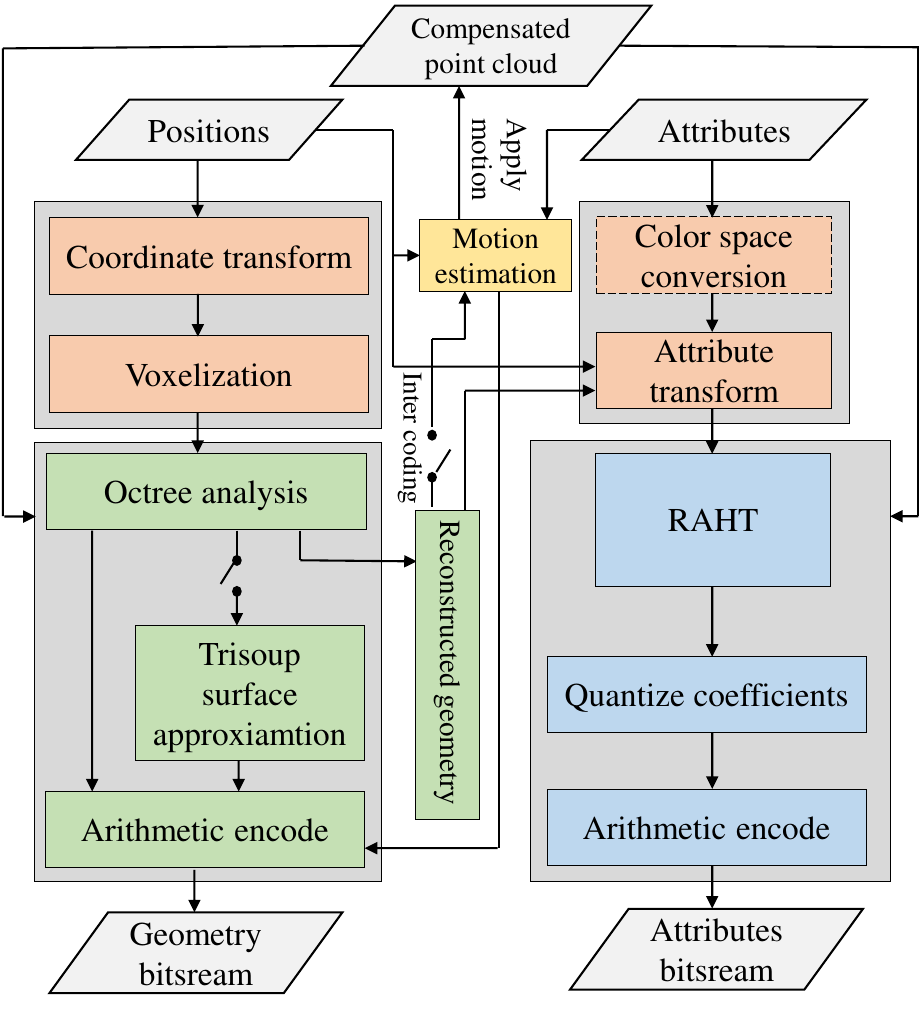}
\caption{Encoding framework of GeS-TM. The modules within the dotted boxes are optional to perform.}
\label{fig1}
\end{figure}

As shown in Fig.\ref{fig1}, in GeS-TM, the geometry and attributes are compressed separately. Initially, coordinate conversion and voxelization are performed to convert point positions into non-negative integers. The encoding process begins with geometry encoding by using either trisoup \cite{ref9} or octree \cite{ref10} to generate geometry bitstream. The decoded geometry is then used to encode the attributes by using the region adaptive hierarchical transform (RAHT) \cite{ref11}, i.e., a hierarchical transform that resembles an adaptive variation of a Haar wavelet. The transformed attributes are subsequently predicted, quantized and arithmetically encoded to generate the attribute bitstream. Additionally, as GeS-TM is primarily designed for dense dynamic point clouds, it also incorporates inter-octree coding with motion compensation \cite{ref12} to enhance the compression efficiency \cite{ref13}.

Attribute quantization inevitably introduces distortion, especially when the quantization parameter (QP) is large. To improve the coding efficiency, quality enhancement for the reconstructed point cloud is an efficient way. Besides, for dynamic point cloud compression with inter prediction, the distortion of the reference frame can also propagate to subsequent frames, causing distortion accumulation \cite{ref14}. Therefore, improving the quality of a reference frame (e.g., the first frame) can enhance the quality of the subsequent frames as well, and thus further improve the coding efficiency. Accordingly, we make the following contributions:
\begin{itemize}

\item We first analyze the feasibility of Wiener filter for the reconstructed point clouds and propose a basic quality enhancement method for color attributes based on Wiener filter, namely BWF. 

\item To improve the coding efficiency, we further propose a coefficients inheritance-based Wiener filter (CIWF) for inter-coded frames and a variance classification-based Wiener filter (VCWF) for Luma component.

\item To reduce the computational complexity, we also propose a Morton code-based fast \emph{k} nearest neighbor search method (M-KNN) for fast derivation of filter coefficients.

\item Finally, by integrating the proposed Wiener filter into the latest G-PCC encoder, significant coding gain can be achieved. Specifically, for Luma, Cb, and Cr components, respectively, average Bjontegaard delta rates of -6.1\%, -7.3\%, and -8.0\% can be achieved under the condition of lossless-geometry-lossy-attributes configuration.

\end{itemize}

The remainder of this paper is organized as follows. In Section \ref{sec:related_work}, we briefly review related work about point cloud quality enhancement. In Section \ref{sec:statistical_analysis}, we analyze the statistical characteristic of reconstructed point clouds, based on which, we describe the proposed method in detail in Section \ref{sec:Proposed method}. After that, in Section \ref{sec:Experimental_results}, we give extensive experimental results to illustrate the advancement of the proposed method. Finally, we conclude this paper and discuss future work in Section \ref{sec:Conclusion}.

\section{Related work}
\label{sec:related_work}
In recent years, technologies for point cloud compression have advanced rapidly, and a variety of quality enhancement methods for compressed point clouds also emerged. Most existing methods, however, focus on static point clouds. In this section, we first give an overview of quality enhancement methods for static point clouds, which can be roughly divided into traditional methods and neural network-based methods. Subsequently, a brief overview of quality enhancement methods for dynamic point clouds is given. 

\subsection{Quality Enhancement for Static Point Clouds}
Traditional quality enhancement methods mainly include Kalman filter-based methods \cite{ref15}, Wiener filter-based methods \cite{ref16,ref17}, and graph signal processing-based methods \cite{ref19,ref20,ref21,ref22,ref23,ref24}. In the previous research, our team \cite{ref15} proposed a Kalman filter-based method to improve not only the reconstruction quality but also the prediction accuracy in the compression pipeline. As Kalman filter is very sensitive to the stationary of signals, it is only efficient for the Chroma red difference and Chroma blue difference, i.e., Cr and Cb, components. For Luma component, its performance is trivial. Therefore, in \cite{ref16} and \cite{ref17}, we developed a Wiener filter-based quality enhancement method for both G-PCC and V-PCC to mitigate coding distortion of color attributes. However, the time complexity is extremely high due to the high complexity of KNN for nearest neighbor search, and the performance of Luma component is also far less than that of Chroma components. Besides, Lin \emph{et al.} \cite{ref18} proposed using median and bilateral filters to improve the color quality of point clouds, which is achieved by assigning the median or weighted average of the nearest neighbor attributes to the current point. This method is more effective for some specific noises, such as salt and pepper noise. 

Since point clouds can naturally be represented as a graph structure, graph signal processing (GSP) has emerged as an effective approach for enhancing point cloud quality. Yamamoto \emph{et al.} \cite{ref19} proposed a deblurring algorithm for point cloud attributes, extending the multi-Wiener SURE-LET deconvolution method \cite{ref20}—typically used for image processing—into the graph spectral domain. This algorithm employs a Wiener-like filter followed by sub-band decomposition and thresholding to improve the quality of deblurred point clouds. However, transforming high-dimensional signals into the graph spectral domain is computationally complex, and the algorithm's performance heavily depends on the choice of graph structure and parameters. Dinesh \emph{et al.} \cite{ref21} introduced a smoothing algorithm known as graph Laplacian regularization, which is applied to enhance the color quality of point clouds. This method functions essentially as a low-pass filter in the graph Fourier domain. However, the graph construction is sensitive to noise, which can affect both color and geometry. To address this limitation, Watanabe \emph{et al.} \cite{ref22} proposed a graph construction method using 3D patch-based similarity. In this approach, patches around two points are compared using isomorphic interpolated 3D patches from neighboring points, making the constructed graphs less susceptible to noise. While this method is effective for additive white Gaussian noise, its applicability to other noise types, such as multiplicative noise, has not been addressed. Additionally, there are quality enhancement methods that exploit the relationship between geometry and color. Irfan and Magli \cite{ref23,ref24} presented methods that integrates both geometry and color for quality enhancement. They constructed a graph with edges weighted by both color and geometry, and then applied convex optimization to achieve effective quality enhancement for both. While the GSP-based methods show promising results, their time complexity and difficulty in constructing an appropriate graph remains a significant challenge.

Recently, neural network-based approaches for point cloud quality enhancement have also been investigated. Tao \emph{et al.} \cite{ref25} proposed a multi-view projection based joint geometry and color hole repairing method for point clouds reconstructed from G-PCC, significantly improving reconstruction quality. While the proposed method effectively detects and repairs holes, its performance does rely on the accuracy of the multi-view projection. Sheng \emph{et al.} \cite{ref26} proposed a multi-scale graph attention network (MS-GAT) to remove the artifacts in point cloud attributes compressed by G-PCC. However, the computational complexity is high due to the Chebyshev graph convolution. Moreover, our team proposed a 3D patch-based graph convolution neural network, namely GQE-Net \cite{ref27}, and a 2D patch-based U-Net tailored for small scale images, namely SSIU-Net \cite{ref28}, to restore the color attributes of distorted point clouds efficiently. Although the neural network-based methods achieve significant performance, their complexity is still very high and is not suitable for integration into G-PCC for real-time application.

\subsection{Quality Enhancement for Dynamic Point Clouds}
Despite these advancements, research on attribute quality enhancement for dynamic point clouds remains limited. Most studies focus primarily on quality enhancement for geometry. For instance, Schoenenberger \emph{et al.} \cite{ref29} briefly discussed how graph-based quality enhancement methods for static point clouds can be adapted for dynamic point clouds. Hu \emph{et al.} \cite{ref30} leveraged the temporal consistency of surface patches corresponding to the same underlying manifold to enhance the geometry quality for dynamic point clouds by utilizing a spatial-temporal graph representation. However, the reliance on accurate estimation of surface normal could be a potential bottleneck, as noise in the input point cloud can adversely affect the quality of these estimates. Yan \emph{et al.} \cite{ref31} introduced a novel quality enhancement method tailored for dynamic point clouds, aiming at enhancing the perceptual capabilities of LiDAR in snowy conditions for autonomous driving. Specifically, it utilizes a time outlier removal filter combined with an entropy weighted method for quality enhancement, showing promising performance. Hong \emph{et al.} \cite{ref32} proposed an efficient vertex domain graph-based low-pass filter to remove noise. It can eliminate high frequency components of the reconstructed reference point cloud frame and therefore improve inter prediction accuracy when there is inaccurate motion during compression of dynamic point clouds. However, the complexity and difficulty of constructing an appropriate graph structure is very challenging. 

Different from existing methods, this paper focuses on attribute quality enhancement for dynamic point clouds by leveraging Wiener filter in a sophisticated manner. The goal is to improve the rate-distortion (RD) performance of G-PCC while maintaining affordable computational complexity.

\section{Statistical Analysis}
\label{sec:statistical_analysis}

\begin{figure}[!t]
\centering
\includegraphics[width=0.47\textwidth]{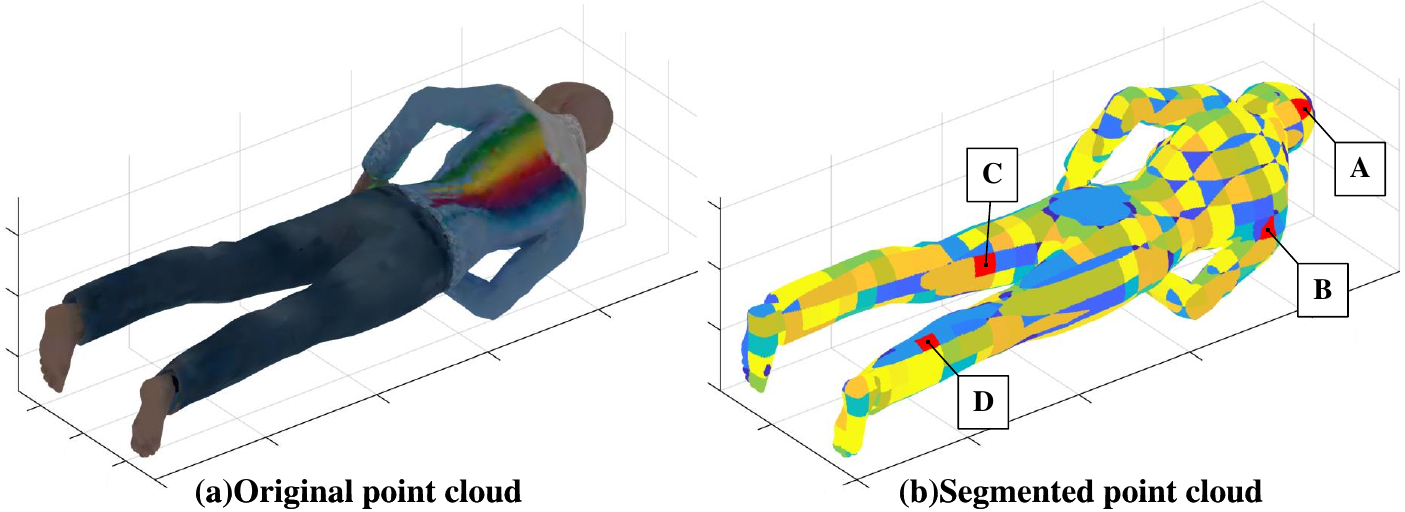}
\caption{Results of octree segmentation, where the octree structure is divided into six layers. The blocks, marked in red and labeled as A, B, C, and D, are selected for local stationary analysis.}
\label{fig2}
\end{figure}

For statistical stationary signals, Wiener filter is an optimal linear filter in terms of minimizing the mean squared error (MSE) between the reconstructed signal and the original signal \cite{ref33}, which was widely used in image quality enhancement. The effectiveness of Wiener filter depends on the correctness of the wide-sense stationary (WSS) \cite{ref34} assumption of the input signal. Specifically, in the spatial domain, WSS assumption requires the statistical properties, such as the mean and autocovariance functions, of an input signal, remain constant in a spatial region \cite{ref35}:
\begin{itemize}
\item The mean of the input signal is near the same at different spatial locations.
\item The autocovariance function of the signal depends only on the spatial distance between two positions, and is independent of the specific locations. 
\end{itemize}

Suppose a spatial sequence $\left\{\bm{x} _{i} \right\} $, for all spatial locations \emph{i} and any spatial distance \emph{n}, the autocovariance of $\bm{x} _{i}$ and $\bm{x} _{i-n}$ is $\sigma \left ( \bm{x} _{i},\bm{x} _{i-n}  \right )=\gamma _{n}$, where $\gamma _{n}$ is independent of the location \emph{i} and only depends on the distance \emph{n}. Specifically, when the spatial distance \emph{n} is 0, the autocovariance formula is expressed as $\sigma \left ( \bm{x} _{i},\bm{x} _{i}  \right )=\gamma _{0}$, indicating that the variance of $\bm{x} _{i}$ should equal to a constant $\gamma _{0}$ that is independent of spatial location.

\begin{figure}[!t]
\centering
\includegraphics[width=0.49\textwidth]{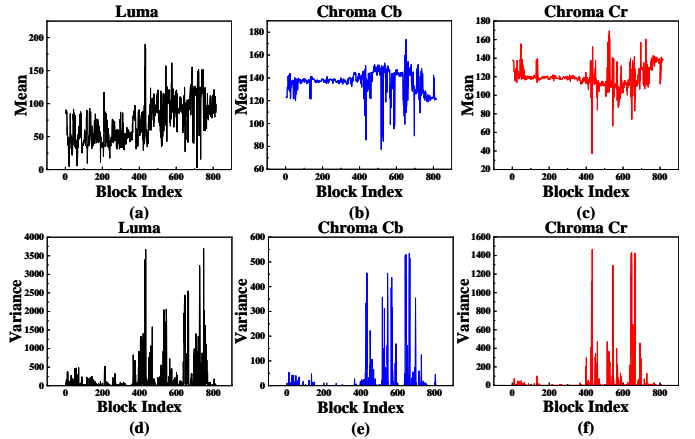}
\caption{Mean and variance distribution of the 816 blocks.}
\label{fig3}
\end{figure}

To check the WSS assumption of color components, we select one frame “\emph{queen\_frame\_1.ply}” from a dense point cloud “\emph{queen}”, as shown in Fig. \ref{fig2}(a), which is encoded and reconstructed with a quantization parameter (QP) of 46 under the condition of lossless-geometry-lossy-attributes configuration. The reconstructed point cloud is then divided into 816 3D blocks using octree segmentation with a depth of 6, as shown in Fig. \ref{fig2}(b). The means and variances for the color (Luma, Chroma Cb, and Chroma Cr) components of the 816 blocks are shown in Fig. \ref{fig3}. We can see that the means and variances are not consistent at different regions, indicating that WSS assumption does not hold for the whole point cloud.

\begin{figure}[!t]
\centering
\includegraphics[width=0.485\textwidth]{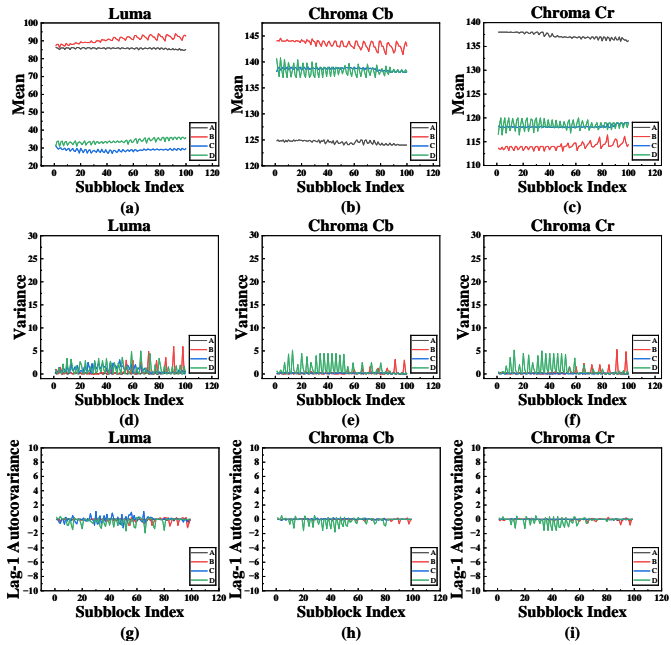}
\caption{Mean, variance and autocovariance ($\gamma _{1}$) distribution of subblocks in block A, B, C and D.}
\label{fig4}
\end{figure}

Furthermore, we selected four blocks from all the segmented blocks, as shown in Fig. \ref{fig2}(b), to test the WSS assumption in local regions. The points in each block are sorted based on their Morton code \cite{ref36} and are divided into 100 subblocks. The means and variances of these subblocks, as well as the autocovariances ($\gamma _{1}$) between two neighboring subblocks are shown in Fig. \ref{fig4}. We can see that the means, variances and autocovariances ($\gamma _{1}$) do not fluctuate too much, indicating that the WSS holds locally.

Based on the above analysis, it is evident that the color components of the reconstructed point cloud are not consistent globally. However, they remain stable locally which is suitable for Wiener filter. Specifically, in the proposed method, the attribute of the current point is filtered by its \emph{k}-1 nearest neighbors and itself. We can also see, in Fig. \ref{fig3} and Fig. \ref{fig4}, that the stationary of Luma component is not as good as that of the Chroma components. Therefore, Luma component should be considered finely when designing the filter.
\section{Proposed method}
\label{sec:Proposed method}
\subsection{Preliminary of Wiener Filter}
\label{sub-sec:1}
Let the order of the filter be \emph{k}. That is to say, the attribute of the current point will be updated by its \emph{k}-1 nearest neighbors and itself. Assuming that the number of points in the point cloud is \emph{n}, the color attribute of the original point cloud can be represented as three column vectors $\left \{\bm{a} _{1},\bm{a} _{2},\bm{a} _{3} \right \} \in \mathbb{R}^{n\times 1} $. Without loss of generality, under the condition of lossless geometry coding, the color attribute of reconstructed point cloud can also be represented as three column vectors $\left \{ \hat{\bm{a}} _{1},\hat{\bm{a}} _{2},\hat{\bm{a}} _{3} \right \} \in \mathbb{R}^{n\times 1} $. Then the $i^{th}$ color attribute of the \emph{k} nearest neighbors in the reconstructed point cloud can be represented as a matrix $\hat{\bm{P}} _{i}\in \mathbb{R}^{n\times k}$. 

Suppose that the coefficients of Wiener filter are a vector $\bm{h}_{i}\in \mathbb{R}^{n\times 1},i\in \left \{ 1,2,3 \right \}$, we can get the filtered attribute $\tilde{\bm{a}} _{i}\in \mathbb{R}^{n\times 1},i\in \left \{ 1,2,3 \right \}$ by
\begin{equation}
\label{deqn_ex1}
\tilde{\bm{a}} _{i} = \hat{\bm{P} } _{i} \times \bm{h} _{i}.
\end{equation}
Then, we get the error $\bm{y}  \in \mathbb{R}^{n\times 1}$:
\begin{equation}
\label{deqn_ex2}
\bm{y} = \bm{a} _{i} - \tilde{\bm{a} } _{i}.
\end{equation}
The aim of Wiener filter is to find an optimal set of coefficients $\bm{h} ^{opt}$ to minimize the objective function:
\begin{equation}
\label{deqn_ex3}
E(\bm{y} ^{2}) = E[({\bm{a} _{i}- \hat{\bm{P} } _{i}\times \bm{h} _{i}})^2],
\end{equation}
where $E(\cdot )$ denotes the average operation over the elements of a vector. The problem can be solved by setting the derivative of this function about $\bm{h} _{i}$ to 0: 
\begin{equation}
\label{deqn_ex4}
\frac{d(E[({\bm{a} _{i}- \hat{\bm{P} } _{i}\times \bm{h} _{i}})^2])}{d\bm{h} _{i}} = 0,
\end{equation}
i.e.,
\begin{equation}
\label{deqn_ex5}
\hat{\bm{P} } _{i}^{T}  \times \bm{a} _{i} - \hat{\bm{P}  } _{i}^{T}\times \hat{\bm{P}} _{i}\times \bm{h} _{i} = 0.
\end{equation}

As the cross-correlation vector $\bm{c} \in \mathbb{R}^{k\times 1}$ between $\hat{\bm{P} } _{i}$ and $\bm{a} _{i}$ can be represented as
\begin{equation}
\label{deqn_ex6}
\bm{c} = \hat{\bm{P} } _{i}^{T}\times \bm{a} _{i},
\end{equation}
and the autocorrelation matrix $\bm{A}\in \mathbb{R}^{k\times k}$ of $\hat{\bm{P} } _{i}$ can be represented as
\begin{equation}
\label{deqn_ex7}
\bm{A} = \hat{\bm{P}} _{i}^{T}\times \hat{\bm{P}},
\end{equation}
(\ref{deqn_ex5}) can be simplified as
\begin{equation}
\label{deqn_ex8}
\bm{c}  - \bm{A} \times \bm{h} _{i} = \bm{0}
\end{equation}
based on which the optimal coefficients $\bm{h} ^{opt}$ can then be expressed as
\begin{equation}
\label{deqn_ex9}
\bm{h} ^{opt} = \bm{A} ^{-1} \times \bm{c}.
\end{equation}

Building on the above principle of Wiener filter, we can obtain a set of filter coefficients for each reconstructed point cloud by finding \emph{k} nearest neighbors of each point. However, the effectiveness of such a naive Wiener filter may be limited due to the complex color distribution as analyzed in Section \ref{sec:statistical_analysis}. Besides, the \emph{k} nearest neighbor search would significantly increase computational complexity.

To address these challenges, we first propose M-KNN for efficient neighbor retrieval in sub-section \ref{sub-sec:2}. To further enhance the quality, we introduce a basic Wiener filter (BWF) specifically designed for intra-coded frames in sub-section \ref{sub-sec:3}, whose quality enhancement effect can also be accumulated over time. Building upon BWF, in sub-sections \ref{sub-sec:4} and \ref{sub-sec:5}, we present the CIWF and VCWF, respectively, both of which significantly improve coding performance.

\subsection{M-KNN}
\label{sub-sec:2}
KNN \cite{ref37} is a commonly used method to search the nearest neighbors. However, for each point in the point cloud, the distances between it and all the other points need to be calculated, resulting in a time complexity of O(\emph{n}). The complexity increases significantly as the number of points increases, which is unacceptable for a real-time encoder. To reduce the time complexity, we propose M-KNN, which derives the \emph{k}-1 nearest neighbors of a point by using the Morton code offsets stored in a search table, and use an index table to quickly determine whether the nearest position is empty or not. 

\begin{figure}[!t]
\centering
\includegraphics[width=0.28\textwidth]{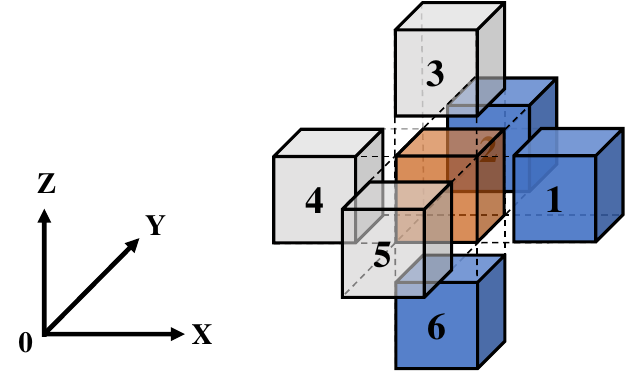}
\caption{Illustration of potential nearest neighbor positions. The brown block denotes the current point position, the blue blocks represent its coplanar nearest neighbor positions which are not empty, while the grey blocks represent its coplanar nearest neighbor positions which are empty.}
\label{fig5}
\end{figure}

Morton codes combine coordinates from multiple dimensions into a one-dimensional number by interleaving the binary representation of each dimension of the coordinates. Assume that the position coordinates of points are represented by three-dimensional Cartesian coordinates $(X,Y,Z)$, and each coordinate is first converted to \emph{N}-bits binary code $(x_N x_{N-1} \dots x_2 x_1)$, $(y_N y_{N-1} \dots y_2 y_1)$, and $(z_N z_{N-1} \dots z_2 z_1)$, where $x_i$, $y_i$, and $z_i$, $i\in \{1,\dots, N\}$, are the corresponding binary codes of $X$, $Y$, and $Z$. Then, the coordinates of $(X,Y,Z)$ can be represented as
\begin{equation}
\begin{aligned}
\label{deqn_ex10}
(X,Y,Z) = & (x_N x_{N-1}\dots x_2 x_1, \\
          &y_N y_{N-1}\dots y_2 y_1, z_N z_{N-1} \dots z_2 z_1).
\end{aligned}
\end{equation}
By interleaving the binary bits of the above coordinates, the Morton code $M$ of $(X,Y,Z)$ can be represented as
\begin{equation}
\label{deqn_ex11}
M=(x_N y_N z_N x_{N-1} y_{N-1} z_{N-1}\dots x_2 y_2 z_2 x_1 y_1 z_1).
\end{equation}
In our approach, we set the filter order as \emph{k} = 7, which requires not only the reconstructed attribute of the current point but also those of its six nearest neighbors. To further reduce the complexity, we focus solely on the nearest co-planar neighbors during M-KNN, as illustrated in Fig. \ref{fig5}.

\begin{table}[]
\caption{Search table that stores the offsets of six coplanar nearest neighbors and the current point}
\centering
\setlength{\tabcolsep}{6.5pt} 
\label{table1}
\begin{tabular}{c|c|c}
\hline
\begin{tabular}[c]{@{}c@{}}Nearest neighbor \\ position index\end{tabular} & \begin{tabular}[c]{@{}c@{}}Cartesian coordinate \\ position difference\end{tabular} & \begin{tabular}[c]{@{}c@{}}Morton code offset \\ (Hexadecimal \\ representation)\end{tabular} \\ \hline
1                                                                          & (1, 0, 0)                                                                           & 0x0000000000000004                                                                            \\
2                                                                          & (0, 1, 0)                                                                           & 0x0000000000000002                                                                            \\
3                                                                          & (0, 0, 1)                                                                           & 0x0000000000000001                                                                            \\
4                                                                          & (-1, 0, 0)                                                                          & 0x4924924924924924                                                                            \\
5                                                                          & (0, -1, 0)                                                                          & 0x2492492492492492                                                                            \\
6                                                                          & (0, 0, -1)                                                                          & 0x9249249249249249                                                                            \\ \hline
\end{tabular}
\end{table}

To implement M-KNN, it is necessary to obtain the Morton code offset \cite{ref36} between the current point position and the nearest neighbor point position. The offset can be derived from the Cartesian coordinate difference between two positions according to the principle of Morton code. For example, as illustrated in Fig. \ref{fig5}, after voxelization, the Cartesian coordinate difference between the point position labeled as 1 and the current point position is (1, 0, 0), while the Morton code offset is 4 which is written as a 64-bit integer 0x0000000000000004 in G-PCC. For the point position labeled as 6, the Cartesian position difference between it and the current point position is (0, 0, -1), while the Morton code offset is 0x9249249249249249 (i.e., the complement of -1). Similarly, the Morton code offsets between the current point position and its six coplanar neighbor positions are calculated and stored in Table \ref{table1}. 

Therefore, the Morton code of the nearest neighbor position can be obtained by adding the Morton code of the current point position and the corresponding offset, as shown in (\ref{deqn_ex12}), 
\begin{equation}
\begin{aligned}
\label{deqn_ex12}
&Morton_{target} = ((a \ AND \ m) + (b \ AND \ m)) \\&\ OR \ ((a \ AND \ (m\ll 1)) + (b \ AND \ (m\ll 1))\\&\ OR \ ((a \ AND \ (m\ll 2)) + (b \ AND \ (m\ll 2)),
\end{aligned}
\end{equation}
where $\ll$ is the left shift operation, $a$ is the Morton code of the current point and $b$ is the corresponding offset, $m$ is a predefined mask which is used to convert a Morton code to Cartesian coordinate and is predefined as a 64 - bit binary number 0x9249249249249249. The AND operation in (\ref{deqn_ex12}) is used to extract the binary code of X, Y, or Z. Taking the Morton code $M$ in (\ref{deqn_ex11}) as an example, to extract the binary code of the Z coordinate, it is only necessary to perform the AND operation between $M$ and $m$, i.e.,
\begin{equation}
\label{deqn_ex13}
M \ AND \ m=(0 0 z_N 0 0 z_{N-1}\dots 0 0 z_2 0 0 z_1).
\end{equation}
Similarly, $m \ll 1$ and $m \ll 2$ can be used to extract the binary codes of Y and X coordinates. The OR operation in (\ref{deqn_ex12}) is used to concatenate three binary codes to obtain a Morton code. 

\begin{figure}[!t]
\centering
\includegraphics[width=0.4\textwidth]{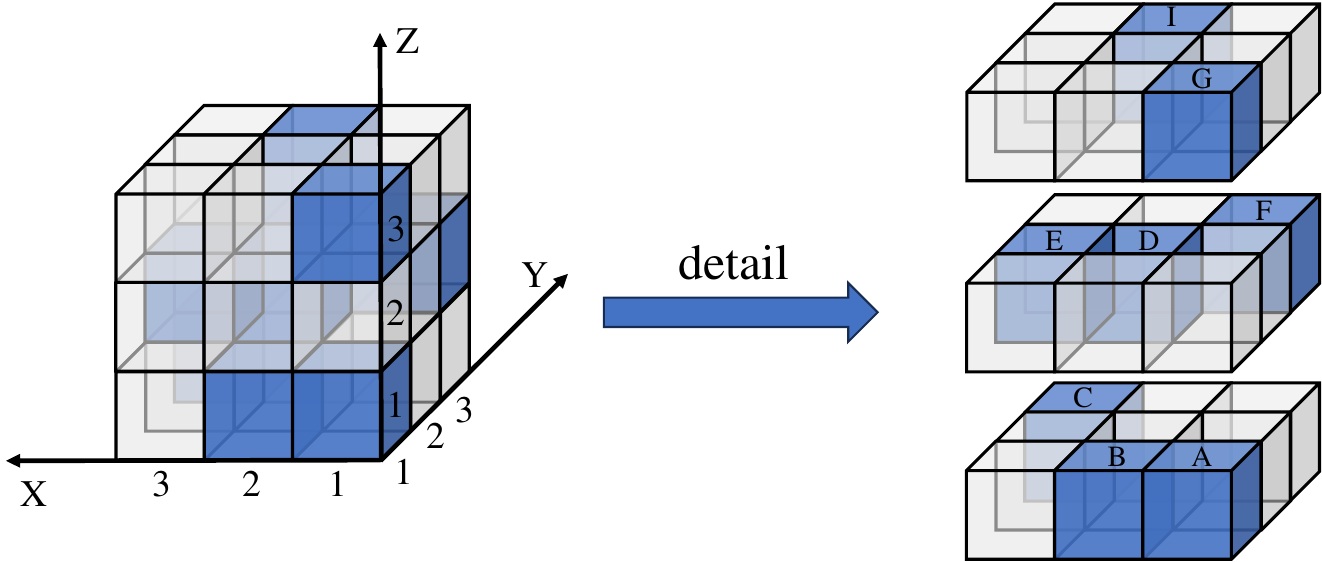}
\caption{Illustration of a 3×3×3 point cloud where blue blocks denote non-empty positions and the others denote empty positions. The points serial number at the non-empty positions are denoted as \{A, B, \dots, I\}, and the corresponding Morton codes are represented as \{M111, M211, \dots, M233\}.}
\label{fig6}
\end{figure}

\begin{table}[]
\caption{Index table based on the 3×3×3 point cloud shown in Fig. \ref{fig6}}
\centering
\setlength{\tabcolsep}{8pt} 
\label{table2}
\begin{tabular}{c|c|c|c|c|c}
\hline
\begin{tabular}[c]{@{}c@{}}Index \\ entry\end{tabular} & \begin{tabular}[c]{@{}c@{}}Index \\ value\end{tabular} & \begin{tabular}[c]{@{}c@{}}Index \\ entry\end{tabular} & \begin{tabular}[c]{@{}c@{}}Index \\ value\end{tabular} & \begin{tabular}[c]{@{}c@{}}Index \\ entry\end{tabular} & \begin{tabular}[c]{@{}c@{}}Index \\ value\end{tabular} \\ \hline
M111        & A           & M112        & null        & M113        & G           \\
M121        & null        & M122        & null        & M123        & null        \\
M131        & null        & M132        & F           & M133        & null        \\
M211        & B           & M212        & null        & M213        & null        \\
M221        & null        & M222        & D           & M223        & null        \\
M231        & null        & M232        & null        & M233        & I           \\
M311        & null        & M312        & null        & M313        & null        \\
M321        & null        & M322        & E           & M323        & null        \\
M331        & C           & M332        & null        & M333        & null        \\ \hline
\end{tabular}
\end{table}

For each neighbor position, it is also necessary to determine whether it contains a point, which can be done by scanning the Morton codes of all points in the point cloud until one point matches the Morton code of the nearest neighbor position. This scanning procedure can be done by using a variety of methods, such as linear search (average time complexity O(\emph{n})) \cite{ref38} or binary search (average time complexity O(\emph{log n}))\cite{ref38}. If the target point cannot be found after traversing the point cloud, the scanning time is wasted. To reduce the time complexity, we only need to traverse the points in the point cloud at the beginning of the encoder and build an index table, as shown in Table \ref{table2} and Fig. \ref{fig6}, to directly index whether there are points in the nearest neighbor, and the average time complexity can be reduced to O(1).

The rule of constructing the index table is as follows. First, an empty index table is created to cover the entire point cloud, where each entry in the table corresponds to the Morton code of a position. Then, points in the point cloud are traversed. For each occupied position, the corresponding indices are set to the serial number (i.e., from A to I, as shown in Fig. \ref{fig6}) of the point, while for empty positions, the indices are set to null, as shown Table \ref{table2}. In practical, using a single index table for the entire point cloud can lead to inefficient usage of memory, as there are so many points. To address this, the input point cloud can be divided into multiple slices \cite{ref39}. By using the index table, we can efficiently check whether a nearest neighbor position is occupied by directly querying the Morton code of that position. During the nearest neighbor search, if a position is empty, a “\emph{virtual attribute}” that is set as the attribute of the current point is assigned to that location.

\begin{algorithm}[h] 
\caption{M-KNN Algorithm}
\label{alg:alg1}
\textbf{Input:} 
$\left \{ P_i, i \in [1, U] \right \} $: points in the input point cloud;
$\left \{ I \right \} $: Index table;
$\left \{ S_j, j \in [1, K] \right \} $: offsets of each point in the search table;
$U$: number of points;
$K$: number of nearest neighbors.
\\\textbf{Output:}
$\left \{ V_{P_i}, i \in [1, U] \right \} $: the set of nearest neighbor attributes of $P_i$.
\begin{algorithmic}[1]
\State $\left \{ P_i \right \} $ is sorted in ascending order based on their Morton codes.
\For{$i = 1$ \textbf{to} $U$}
    \State Calculate the Morton code of $P_i$.
    \For{$j = 1$ \textbf{to} $K$}
        \State Calculate the Morton code of the coplanar adjacent position $M_{ij}$ by traversing $S_j$.
        \State Index in $I$ using $M_{ij}$, get the index value $ \emph{index}_i$.
        \If{$ \emph{index}_i =  \emph{null}$}
            \State Record the attribute of the indexed point as $V_{P_i}$, a ``virtual attribute''.
        \Else
            \State Record the attribute of the point whose sequence number is $\text{index}_i$ to $V_{P_i}$.
        \EndIf
    \EndFor
\EndFor
\end{algorithmic}
\end{algorithm}

In summary, the implementation of the proposed M-KNN is outlined in Algorithm \ref{alg:alg1}. First, points are sorted in ascending order based on their Morton codes. Next, for each point, the Morton codes of its coplanar adjacent positions are computed by traversing the offsets in the search table, e.g., Table \ref{table1}. Then, the calculated Morton codes of the coplanar adjacencies are used to index the corresponding positions in the index table, e.g. Table \ref{table2}. If a position is occupied, the attribute of the point at that position is recorded as a neighbor attribute; if the position is empty, the attribute of the current point is recorded as the neighbor attribute. 
\subsection{BWF for Dynamic Point Clouds}
\label{sub-sec:3}

\begin{figure}[!t]
\centering
\includegraphics[width=0.49\textwidth]{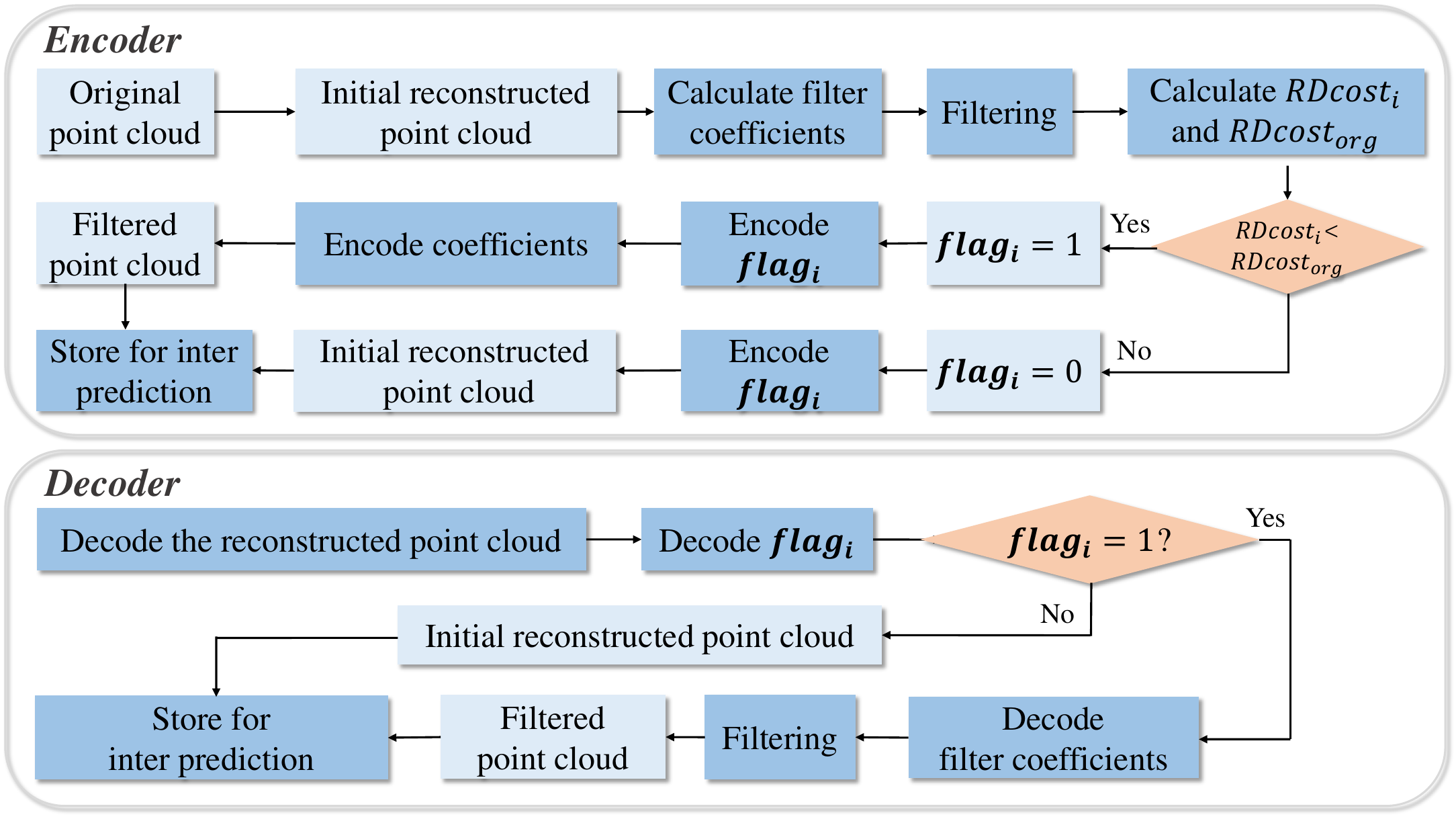} 
\caption{Encoding and decoding flowcharts of BWF for the $i^{th}$ color component. $RDcost_{i}$ is the RD cost by implementing the BWF for the $i^{th}$ color component, while $RDcost_{org}$ is RD cost without BWF.}
\label{fig7}
\end{figure}

According to the analysis in Appendix, the quality improvement in one frame can also benefit the subsequent frames. Accordingly, we propose BWF for dynamic point clouds, as illustrated in Fig. \ref{fig7}, in which all the frames will be filtered by their corresponding filter coefficients, to improve the coding efficiency. 

As the filter coefficients need to be encoded into the bitstream, to determine whether the Wiener filter is effective or not in terms of RD performance, the following RD cost \cite{ref40} is calculated:
\begin{equation}
\label{deqn_ex17}
RDcost_{i} = D_{i} + \lambda \times R_{i},
\end{equation}
where $D_{i}$ represents the sum of squared error (SSE) of the $i^{th}$  color component between the reconstructed point cloud and the original point cloud, $R_{i}$ is the coding bits (including the filter coefficients) of the $i^{th}$ color component, $RDcost_{i}$ is the total RD cost of the $i^{th}$ color component, and $\lambda$ is the Lagrange multiplier, which is related to QP \cite{ref41}:
\begin{equation}
\label{deqn_ex18}
\lambda = 0.85 \times 2^{\frac{QP-12}{3}}.
\end{equation}

The smaller the $RDcost_{i}$, the better the encoding performance. By comparing the RD cost before and after the BWF, we can determine whether to perform it or not. For each frame in the dynamic point cloud, a 1-bit flag $flag_{i}$, $i\in \left \{ 1,2,3 \right \} $, is also assigned to each color component and encoded into the bitstream. Specifically, if BWF improves the RD performance for the $i^{th}$ color component, $flag_{i}$ is set to 1, and the corresponding filter coefficients are encoded into the bitstream. Otherwise, $flag_{i}$ is set to 0, and the filter coefficients are omitted.

The decoder, as illustrated in Fig. \ref{fig7}, parses the $flag_{i}$, $i\in \left \{ 1,2,3 \right \} $, followed by the filter coefficients. Then, if $flag_{i}$ is 1, the BWF will be performed to improve the quality of the current frame that is then used as a reference to decode the subsequent frames. This procedure is repeated for each frame, until all frames are decoded. 
\subsection{CIWF for Saving Filter Coefficients}
\label{sub-sec:4}
\begin{figure}[!t]
\centering
\includegraphics[width=0.48\textwidth]{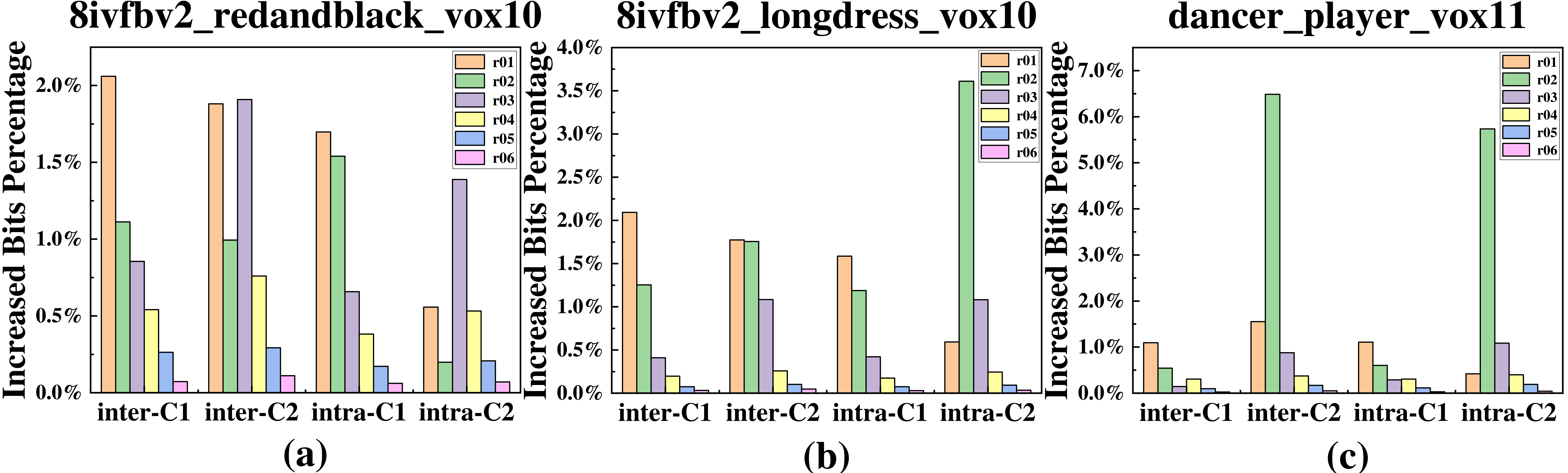} 
\caption{The percentages of of coding bits increment caused by BWF. The geometry is compressed by octree encoding while the attributes are compressed by RAHT, C1 and C2 represent lossless-geometry-lossy-attributes and lossy-geometry-lossy-attributes configurations, respectively.}
\label{fig8}
\end{figure}

Although BWF can significantly improve the coding efficiency, the storage of the filter coefficients consumes a lot of bits, which limits the RD performance, especially for low bitrate coding configuration. Fig. \ref{fig8} shows the percentage of bit increase induced by BWF at six QPs (corresponding to six coding bitrates: r01, r02, r03, r04, r05, and r06, from low to high) \cite{ref42}, under the octree geometry coding configuration specified by the common test condition of GeS-TM, for the point clouds “\emph{8ivfbv2\_redandblack\_vox10}”, “\emph{8ivfbv2\_longdress\_vox10}”, and “\emph{dancer\_player\_vox11}” \cite{ref42}.

\begin{figure}[!t]
\centering
\includegraphics[width=0.48\textwidth]{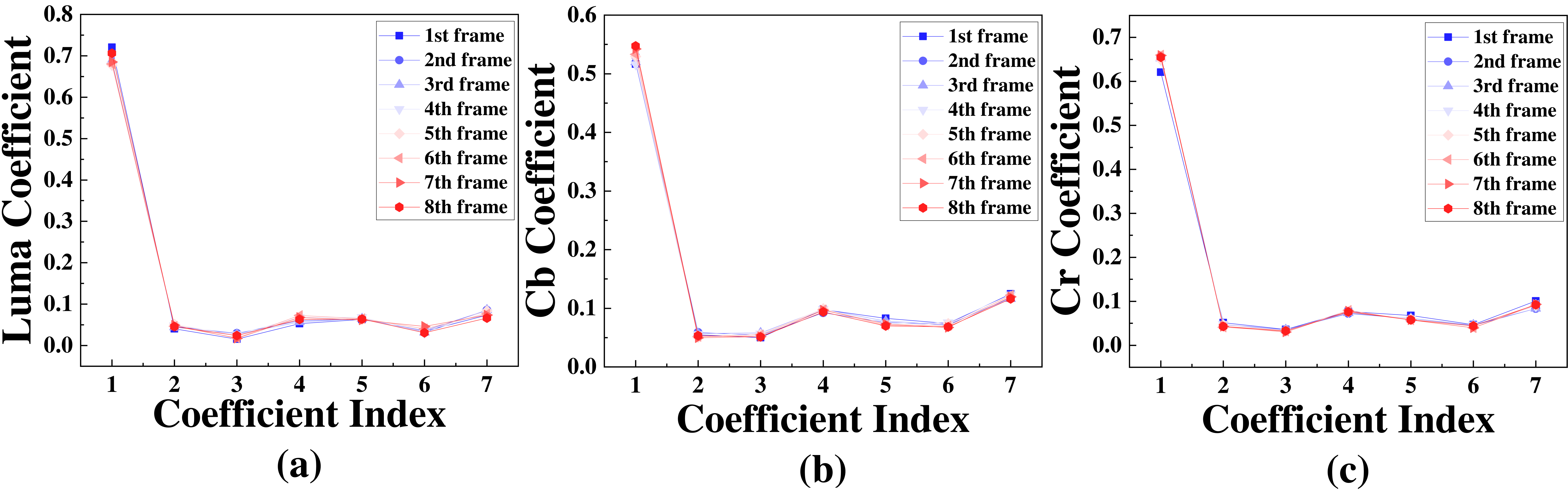} 
\caption{Filter coefficients of eight consecutive frames for the dynamic point cloud “\emph{8ivfbv2\_loot\_vox10}”.}
\label{fig9}
\end{figure}

As the content of adjacent frames in dynamic point cloud are similar, filter coefficients may also be similar. To validate this assumption, we selected eight adjacent frames from the dynamic point cloud “\emph{8ivfbv2\_loot\_vox10}” to calculate the filter coefficients of their color components, as shown in Fig. \ref{fig9}. We can see that the filter coefficients of adjacent frames are very similar. By sharing the same filter coefficients to these adjacent frames, the bitrates consumed by the filter coefficients can be reduced greatly.

\begin{figure}[!t]
\centering
\includegraphics[width=0.49\textwidth]{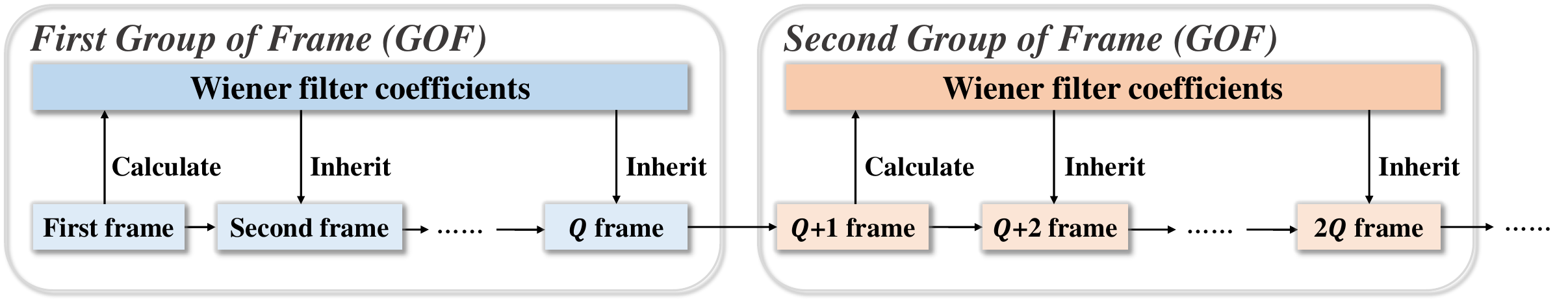} 
\caption{Inheritance of Wiener filter coefficients. }
\label{fig10}
\end{figure}

To reduce the bit consumption of filter coefficients, it is evident to share the same filter coefficients across consecutive frames. Building on the inter-prediction structure of the current GeS-TM, we compute and encode the filter coefficients only for the first frame in each group of frames (GOF), which consists of \emph{G} frames (e.g., 8 frames), as shown in Fig. \ref{fig10}. To maintain RD performance, we also incorporate RD optimization (RDO) at the encoder for each frame to decide whether or not to apply the Wiener filter, as described in sub-section \ref{sub-sec:3}.

\begin{figure}[]
\centering
\includegraphics[width=0.475\textwidth]{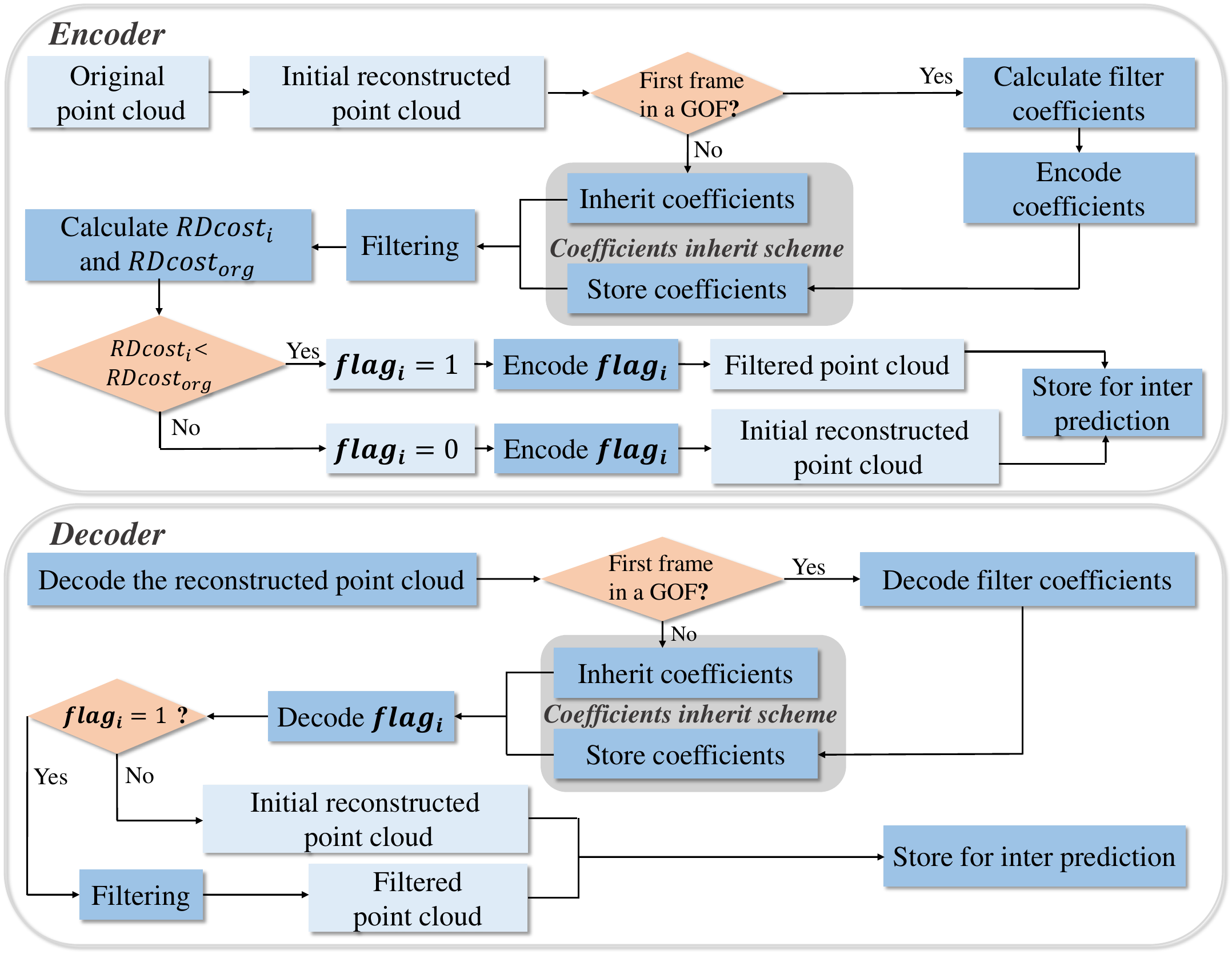} 
\caption{Encoding and decoding flowcharts for CIWF of the $i^{th}$ color component. $RDcost_{i}$ is the RD cost by implementing the CIWF for the $i^{th}$ color component, while $RDcost_{org}$ is RD cost without CIWF.}
\label{fig11}
\end{figure}

The encoder and decoder flowcharts are shown in Fig. \ref{fig11}. To implement the CIWF, during encoding, the filter coefficients for the first frame in a GOF are calculated and stored in a buffer. For the subsequent frames within the GOF, the coefficients are retrieved from the buffer, eliminating the recalculation. At the decoder, the filter coefficients of the first frame in a GOF are decoded from the bitstream and then stored in the buffer. For the remaining frames in the GOF, the decoder can get the coefficients from the buffer for filtering. This approach ensures efficient utilization of filter coefficients while removing redundant computations. 

\subsection{VCWF for Luma}
\label{sub-sec:5}
Statistical analysis in Section \ref{sec:statistical_analysis} indicates that the stationary of the Luma component is lower than that of the Chroma components. Therefore, applying a single set of Luma filter coefficients across the entire frame may not yield the optimal result. As the stationary of Luma component is mainly deteriorated by attribute's fluctuation, we propose to divide the Luma component of points into different categories in terms of the variance between the current point and its \emph{k}-1 nearest neighbors and calculate filter coefficients for each category separately.

Assuming that the Luma component of the current point is $L_{0}$ while the Luma component of its $i^{th}$ nearest point is $L_{i}$,$i\in \left \{ 0,\dots,\emph{k}-1 \right \}$, the average of them are
\begin{equation}
\label{deqn_ex19}
\mu = \frac{L_{0}+ {\textstyle \sum_{i=1}^{k-1}L_{i}}  }{k},
\end{equation}
and their variance is

\begin{equation}
\begin{aligned}
\label{deqn_ex20}
V &= \frac{1}{k}  {\textstyle \sum_{i=0}^{k-1}\left ( L_{i} - \mu  \right )^2 } \\
  &= \frac{1}{k} \left [ \left ( L_{0} - \mu  \right )^2 +  {\textstyle \sum_{i=1}^{k-1}\left ( L_{i} - \mu   \right )^2 } \right ].
\end{aligned}
\end{equation}

\begin{figure}[]
\centering
\includegraphics[width=0.475\textwidth]{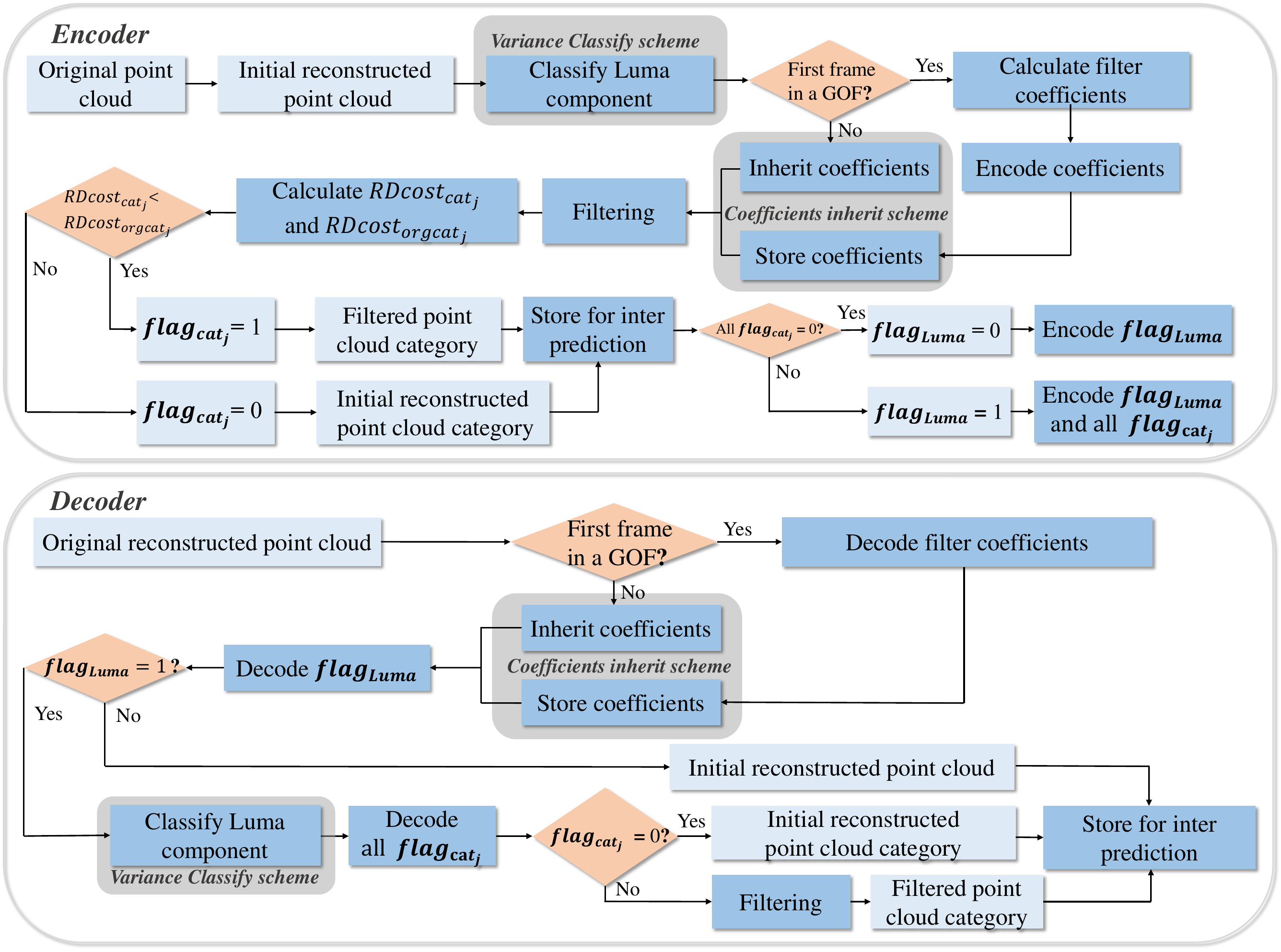} 
\caption{Encoding and decoding flowcharts of VCWF for Luma. $flag_{cat_{j}}$ is the RD cost by implementing the VCWF for the $j^{th}$ class of Luma component, while $flag_{orgcat_{j}}$ is the RD cost without VCWF for the $j^{th}$ class of Luma component.}
\label{fig12}
\end{figure}

Based on the variances calculated from the \emph{k}-1 nearest neighbors of each point, we classify the points into five categories, enabling the computation of five distinct sets of filter coefficients for the Luma component. The classification thresholds are set as 10, 20, 40 and 60 empirically. The flowchart of the process is presented in Fig. \ref{fig12}. For each point in the first frame of a GOF, the variances of all the points are first computed and categorized. Then, the filter coefficients for each category are calculated and encoded, followed by an RDO to determine whether to apply the Wiener filter, indicated by $flag_{cat_{j}}$, $j\in \left \{ 1,\dots,5 \right \}$. Next, the same variance-based classification is applied to the other frames in the GOF. For each category in the successive frames, the corresponding filter coefficients calculated in the first frame are applied directly, and RDO is also used to determine whether to perform Wiener filter for each category. Accordingly, for a GOF, we encode five sets of filter coefficients for the Luma component and two sets for the Chroma components. If none of the five categories need filtering, i.e., the Luma component does not need to be filtered, $flag_{Luma}$  is set to 0 while $flag_{cat_{j}}$ does not need to be encoded in the bitstream.

The decoder flowchart is shown in Fig. \ref{fig12}. For Luma component of the first frame in each GOF, five groups of filter coefficients are decoded from the bitstream and stored in the buffer. For the rest of frames in each GOF, filter coefficients are read from the buffer directly. After that, $flag_{Luma}$ is parsed from the bitstream. If $flag_{Luma}$ = 0, VCWF are not required. If $flag_{Luma}$ = 1, for each point in this frame, variance-based classification is carried out, and $flag_{cat_{j}}$, $j\in \left \{ 1,\dots,5 \right \}$, which indicates whether the $j^{th}$ category needs filtering, is decoded from the bitstream, then the corresponding filter coefficients are used to filter the $j^{th}$ category based on the instruction of $flag_{cat_{j}}$.


\section{Experimental results}
\label{sec:Experimental_results}

\begin{table*}[]
\caption{RD performance comparison between BWF-M and GeS-TMv7.0-rc2 under C1 and C2 condition}
\centering
\label{table3}
\setlength{\tabcolsep}{6pt}  
\renewcommand{\arraystretch}{0.97}
\begin{tabular}{cllllllllllll}
\hline
                        & \multicolumn{12}{c}{End-to-End BD AttrRate (\%)}                                                                                                                                                                                                                                                                                                                                                                                                                                                                     \\
                        & \multicolumn{3}{c}{Octree-raht-intra-C1}                                                                         & \multicolumn{3}{c}{Octree-raht-inter-C1}                                                                         & \multicolumn{3}{c}{Octree-raht-intra-C2}                                                                                                         & \multicolumn{3}{c}{Octree-raht-inter-C2}                                                                                    \\
\multirow{-3}{*}{Class} & \multicolumn{1}{c}{Luma} & \multicolumn{1}{c}{Cb} & \multicolumn{1}{c}{Cr}                                       & \multicolumn{1}{c}{Luma} & \multicolumn{1}{c}{Cb} & \multicolumn{1}{c}{Cr}                                       & \multicolumn{1}{c}{Luma}                & \multicolumn{1}{c}{Cb}                  & \multicolumn{1}{c}{Cr}                                       & \multicolumn{1}{c}{Luma}                & \multicolumn{1}{c}{Cb}                  & \multicolumn{1}{c}{Cr}                  \\ \hline
\rowcolor[HTML]{EFEFEF} 
loot                    & -3.1\%                   & -0.2\%                 & \multicolumn{1}{l|}{\cellcolor[HTML]{EFEFEF}-1.6\%}          & -4.1\%                   & -0.3\%                 & \multicolumn{1}{l|}{\cellcolor[HTML]{EFEFEF}-1.3\%}          & \cellcolor[HTML]{EFEFEF}-0.5\%          & \cellcolor[HTML]{EFEFEF}-0.0\%          & \multicolumn{1}{l|}{\cellcolor[HTML]{EFEFEF}-1.0\%}          & \cellcolor[HTML]{EFEFEF}-0.1\%          & \cellcolor[HTML]{EFEFEF}1.8\%           & \cellcolor[HTML]{EFEFEF}1.5\%           \\
redandblack             & -3.3\%                   & -2.0\%                 & \multicolumn{1}{l|}{-4.4\%}                                  & -3.3\%                   & -0.6\%                 & \multicolumn{1}{l|}{-5.8\%}                                  & -0.6\%                                  & -1.1\%                                  & \multicolumn{1}{l|}{-0.3\%}                                  & 0.4\%                                   & 0.3\%                                   & 0.0\%                                   \\
\rowcolor[HTML]{EFEFEF} 
soldier                 & -3.6\%                   & -0.7\%                 & \multicolumn{1}{l|}{\cellcolor[HTML]{EFEFEF}-0.8\%}          & -4.6\%                   & -5.4\%                 & \multicolumn{1}{l|}{\cellcolor[HTML]{EFEFEF}-5.0\%}          & \cellcolor[HTML]{EFEFEF}-0.2\%          & \cellcolor[HTML]{EFEFEF}0.4\%           & \multicolumn{1}{l|}{\cellcolor[HTML]{EFEFEF}0.5\%}           & \cellcolor[HTML]{EFEFEF}-1.0\%          & \cellcolor[HTML]{EFEFEF}-5.5\%          & \cellcolor[HTML]{EFEFEF}-4.5\%          \\
queen                   & \textbf{-5.1\%}          & \textbf{-11.5\%}       & \multicolumn{1}{l|}{\textbf{-11.8\%}}                        & \textbf{-7.1\%}          & \textbf{-16.7\%}       & \multicolumn{1}{l|}{\textbf{-17.4\%}}                        & \textbf{-2.4\%}                         & \textbf{-7.1\%}                         & \multicolumn{1}{l|}{\textbf{-6.4\%}}                         & \textbf{-7.4\%}                         & \textbf{-14.1\%}                        & \textbf{-13.9\%}                        \\
\rowcolor[HTML]{EFEFEF} 
longdress               & -2.4\%                   & -4.9\%                 & \multicolumn{1}{l|}{\cellcolor[HTML]{EFEFEF}-5.1\%}          & -2.4\%                   & -4.6\%                 & \multicolumn{1}{l|}{\cellcolor[HTML]{EFEFEF}-5.4\%}          & \cellcolor[HTML]{EFEFEF}0.1\%           & \cellcolor[HTML]{EFEFEF}-1.6\%          & \multicolumn{1}{l|}{\cellcolor[HTML]{EFEFEF}-0.4\%}          & \cellcolor[HTML]{EFEFEF}0.2\%           & \cellcolor[HTML]{EFEFEF}-1.2\%          & \cellcolor[HTML]{EFEFEF}-0.5\%          \\
basketball\_player      & -2.4\%                   & -0.5\%                 & \multicolumn{1}{l|}{-1.7\%}                                  & -2.9\%                   & -0.4\%                 & \multicolumn{1}{l|}{-2.0\%}                                  & -0.4\%                                  & 2.1\%                                   & \multicolumn{1}{l|}{0.2\%}                                   & -1.0\%                                  & 2.5\%                                   & 2.1\%                                   \\
\rowcolor[HTML]{EFEFEF} 
dancer\_player          & -2.7\%                   & -0.8\%                 & \multicolumn{1}{l|}{\cellcolor[HTML]{EFEFEF}-1.7\%}          & -3.0\%                   & -0.4\%                 & \multicolumn{1}{l|}{\cellcolor[HTML]{EFEFEF}-0.6\%}          & \cellcolor[HTML]{EFEFEF}0.0\%           & \cellcolor[HTML]{EFEFEF}1.7\%           & \multicolumn{1}{l|}{\cellcolor[HTML]{EFEFEF}0.6\%}           & \cellcolor[HTML]{EFEFEF}-0.3\%          & \cellcolor[HTML]{EFEFEF}2.2\%           & \cellcolor[HTML]{EFEFEF}2.4\%           \\ \hline
Cat2-A average          & -3.8\%                   & -3.6\%                 & \multicolumn{1}{l|}{-4.6\%}                                  & -4.8\%                   & -5.7\%                 & \multicolumn{1}{l|}{-7.4\%}                                  & -0.9\%                                  & -1.9\%                                  & \multicolumn{1}{l|}{-1.8\%}                                  & -2.0\%                                  & -4.4\%                                  & -4.2\%                                  \\
\rowcolor[HTML]{EFEFEF} 
Cat2-B average          & -2.4\%                   & -4.9\%                 & \multicolumn{1}{l|}{\cellcolor[HTML]{EFEFEF}-5.1\%}          & -2.4\%                   & -4.6\%                 & \multicolumn{1}{l|}{\cellcolor[HTML]{EFEFEF}-5.4\%}          & \cellcolor[HTML]{EFEFEF}0.1\%           & \cellcolor[HTML]{EFEFEF}-1.6\%          & \multicolumn{1}{l|}{\cellcolor[HTML]{EFEFEF}-0.4\%}          & \cellcolor[HTML]{EFEFEF}0.2\%           & \cellcolor[HTML]{EFEFEF}-1.2\%          & \cellcolor[HTML]{EFEFEF}-0.5\%          \\
Cat2-C average          & -2.5\%                   & -0.6\%                 & \multicolumn{1}{l|}{-1.7\%}                                  & -2.9\%                   & -0.4\%                 & \multicolumn{1}{l|}{-1.3\%}                                  & -0.2\%                                  & 1.9\%                                   & \multicolumn{1}{l|}{0.4\%}                                   & -0.7\%                                  & 2.3\%                                   & 2.3\%                                   \\
\rowcolor[HTML]{EFEFEF} 
Overall average         & \textbf{-3.2\%}          & \textbf{-3.0\%}        & \multicolumn{1}{l|}{\cellcolor[HTML]{EFEFEF}\textbf{-3.9\%}} & \textbf{-3.9\%}          & \textbf{-4.0\%}        & \multicolumn{1}{l|}{\cellcolor[HTML]{EFEFEF}\textbf{-5.4\%}} & \cellcolor[HTML]{EFEFEF}\textbf{-0.6\%} & \cellcolor[HTML]{EFEFEF}\textbf{-0.8\%} & \multicolumn{1}{l|}{\cellcolor[HTML]{EFEFEF}\textbf{-1.0\%}} & \cellcolor[HTML]{EFEFEF}\textbf{-1.3\%} & \cellcolor[HTML]{EFEFEF}\textbf{-2.0\%} & \cellcolor[HTML]{EFEFEF}\textbf{-1.8\%} \\ \hline
\end{tabular}
\end{table*}

To verify the performance of the proposed method, we implemented it on the latest version of the GeS-TM version 7.0 release candidate 2 (GeS-TMv7.0-rc2) \cite{ref43}, and compared the subjective and objective quality of the reconstructed point clouds. We used the dynamic point cloud dataset provided by MPEG, denoted as Category 2 (Cat2) \cite{ref42} in the common test condition (CTC) \cite{ref42} of GeS-TM, for test. The tested point clouds are further categorized as Cat2-A (including \emph{loot}, \emph{redandblack}, \emph{soldier} and \emph{queen}), Cat2-B (including \emph{longdress}), and Cat2-C (including \emph{basketball\_player} and \emph{dancer\_player}) based on the content complexity, i.e., Cat2-A is the lowest and Cat2-C the highest. The hardware platform is an Intel I7-13790K CPU with 32GB memory, and the software platform is Windows 10 operating system. The experiments were conducted according to the CTC of GeS-TMv7.0, i.e., the geometry was compressed by octree encoding while the attributes were compressed by RAHT. Specifically, we used two test conditions: lossless-geometry-lossy-attributes and lossy-geometry-lossy-attributes (namely C1 and C2 conditions, respectively) and compared the results of both intra-only prediction (namely octree-raht-intra) and inter prediction (namely octree-raht-inter).

\begin{table*}[]
\caption{RD performance comparison between CIWF-M and GeS-TMv7.0-rc2 under C1 and C2 condition}
\centering
\setlength{\tabcolsep}{6pt}  
\renewcommand{\arraystretch}{0.97}
\label{table4}
\begin{tabular}{cllllllllllll}
\hline
                        & \multicolumn{12}{c}{End-to-End BD AttrRate (\%)}                                                                                                                                                                                                                                                                                                                                                                                                                                                                     \\
                        & \multicolumn{3}{c}{Octree-raht-intra-C1}                                                                         & \multicolumn{3}{c}{Octree-raht-inter-C1}                                                                         & \multicolumn{3}{c}{Octree-raht-intra-C2}                                                                                                         & \multicolumn{3}{c}{Octree-raht-inter-C2}                                                                                    \\
\multirow{-3}{*}{Class} & \multicolumn{1}{c}{Luma} & \multicolumn{1}{c}{Cb} & \multicolumn{1}{c}{Cr}                                       & \multicolumn{1}{c}{Luma} & \multicolumn{1}{c}{Cb} & \multicolumn{1}{c}{Cr}                                       & \multicolumn{1}{c}{Luma}                & \multicolumn{1}{c}{Cb}                  & \multicolumn{1}{c}{Cr}                                       & \multicolumn{1}{c}{Luma}                & \multicolumn{1}{c}{Cb}                  & \multicolumn{1}{c}{Cr}                  \\ \hline
\rowcolor[HTML]{EFEFEF} 
loot                    & -3.8\%                   & -2.6\%                 & \multicolumn{1}{l|}{\cellcolor[HTML]{EFEFEF}-4.0\%}          & -5.1\%                   & -6.4\%                 & \multicolumn{1}{l|}{\cellcolor[HTML]{EFEFEF}-6.8\%}          & \cellcolor[HTML]{EFEFEF}-2.2\%          & \cellcolor[HTML]{EFEFEF}-2.4\%          & \multicolumn{1}{l|}{\cellcolor[HTML]{EFEFEF}-4.3\%}          & \cellcolor[HTML]{EFEFEF}-1.5\%          & \cellcolor[HTML]{EFEFEF}-2.3\%          & \cellcolor[HTML]{EFEFEF}-4.3\%          \\
redandblack             & -3.9\%                   & -3.4\%                 & \multicolumn{1}{l|}{-5.0\%}                                  & -4.1\%                   & -2.9\%                 & \multicolumn{1}{l|}{-6.5\%}                                  & -2.7\%                                  & -2.5\%                                  & \multicolumn{1}{l|}{-1.6\%}                                  & -2.3\%                                  & -2.6\%                                  & -1.6\%                                  \\
\rowcolor[HTML]{EFEFEF} 
soldier                 & -4.0\%                   & -1.9\%                 & \multicolumn{1}{l|}{\cellcolor[HTML]{EFEFEF}-2.3\%}          & -2.7\%                   & -6.9\%                 & \multicolumn{1}{l|}{\cellcolor[HTML]{EFEFEF}-7.4\%}          & \cellcolor[HTML]{EFEFEF}-1.6\%          & \cellcolor[HTML]{EFEFEF}-1.7\%          & \multicolumn{1}{l|}{\cellcolor[HTML]{EFEFEF}-2.0\%}          & \cellcolor[HTML]{EFEFEF}0.5\%           & \cellcolor[HTML]{EFEFEF}-7.1\%          & \cellcolor[HTML]{EFEFEF}-6.8\%          \\
queen                   & \textbf{-5.8\%}          & \textbf{-12.2\%}       & \multicolumn{1}{l|}{\textbf{-11.9\%}}                        & \textbf{-8.4\%}          & \textbf{-20.1\%}       & \multicolumn{1}{l|}{\textbf{-19.2\%}}                        & \textbf{-2.7\%}                         & \textbf{-5.9\%}                         & \multicolumn{1}{l|}{\textbf{-4.8\%}}                         & \textbf{-5.3\%}                         & \textbf{-11.5\%}                        & \textbf{-9.0\%}                         \\
\rowcolor[HTML]{EFEFEF} 
longdress               & -2.5\%                   & -5.3\%                 & \multicolumn{1}{l|}{\cellcolor[HTML]{EFEFEF}-5.4\%}          & -2.6\%                   & -5.0\%                 & \multicolumn{1}{l|}{\cellcolor[HTML]{EFEFEF}-5.8\%}          & \cellcolor[HTML]{EFEFEF}-0.7\%          & \cellcolor[HTML]{EFEFEF}-3.5\%          & \multicolumn{1}{l|}{\cellcolor[HTML]{EFEFEF}-2.3\%}          & \cellcolor[HTML]{EFEFEF}-0.7\%          & \cellcolor[HTML]{EFEFEF}-3.7\%          & \cellcolor[HTML]{EFEFEF}-3.0\%          \\
basketball\_player      & -2.7\%                   & -1.0\%                 & \multicolumn{1}{l|}{-2.8\%}                                  & -2.8\%                   & -1.2\%                 & \multicolumn{1}{l|}{-3.7\%}                                  & -2.4\%                                  & -0.2\%                                  & \multicolumn{1}{l|}{-2.2\%}                                  & -3.3\%                                  & -1.2\%                                  & -5.0\%                                  \\
\rowcolor[HTML]{EFEFEF} 
dancer\_player          & -2.9\%                   & -1.4\%                 & \multicolumn{1}{l|}{\cellcolor[HTML]{EFEFEF}-2.8\%}          & -3.3\%                   & -1.6\%                 & \multicolumn{1}{l|}{\cellcolor[HTML]{EFEFEF}-2.3\%}          & \cellcolor[HTML]{EFEFEF}-1.6\%          & \cellcolor[HTML]{EFEFEF}-0.4\%          & \multicolumn{1}{l|}{\cellcolor[HTML]{EFEFEF}-2.1\%}          & \cellcolor[HTML]{EFEFEF}-2.0\%          & \cellcolor[HTML]{EFEFEF}0.2\%           & \cellcolor[HTML]{EFEFEF}-2.2\%          \\ \hline
Cat2-A average          & -3.7\%                   & -5.0\%                 & \multicolumn{1}{l|}{-5.8\%}                                  & -5.1\%                   & -9.1\%                 & \multicolumn{1}{l|}{-10.0\%}                                 & -2.3\%                                  & -3.1\%                                  & \multicolumn{1}{l|}{-3.2\%}                                  & -2.1\%                                  & -5.9\%                                  & -5.4\%                                  \\
\rowcolor[HTML]{EFEFEF} 
Cat2-B average          & -2.5\%                   & -5.3\%                 & \multicolumn{1}{l|}{\cellcolor[HTML]{EFEFEF}-5.4\%}          & -2.6\%                   & -5.0\%                 & \multicolumn{1}{l|}{\cellcolor[HTML]{EFEFEF}-5.8\%}          & \cellcolor[HTML]{EFEFEF}-0.7\%          & \cellcolor[HTML]{EFEFEF}-3.5\%          & \multicolumn{1}{l|}{\cellcolor[HTML]{EFEFEF}-2.3\%}          & \cellcolor[HTML]{EFEFEF}-0.7\%          & \cellcolor[HTML]{EFEFEF}-3.7\%          & \cellcolor[HTML]{EFEFEF}-3.0\%          \\
Cat2-C average          & -2.8\%                   & -1.2\%                 & \multicolumn{1}{l|}{-2.8\%}                                  & -3.1\%                   & -1.4\%                 & \multicolumn{1}{l|}{-3.0\%}                                  & -2.0\%                                  & -0.3\%                                  & \multicolumn{1}{l|}{-2.1\%}                                  & -2.7\%                                  & -0.5\%                                  & -3.6\%                                  \\
\rowcolor[HTML]{EFEFEF} 
Overall average         & \textbf{-3.7\%}          & \textbf{-4.0\%}        & \multicolumn{1}{l|}{\cellcolor[HTML]{EFEFEF}\textbf{-4.9\%}} & \textbf{-4.1\%}          & \textbf{-6.3\%}        & \multicolumn{1}{l|}{\cellcolor[HTML]{EFEFEF}\textbf{-7.4\%}} & \cellcolor[HTML]{EFEFEF}\textbf{-2.0\%} & \cellcolor[HTML]{EFEFEF}\textbf{-2.4\%} & \multicolumn{1}{l|}{\cellcolor[HTML]{EFEFEF}\textbf{-2.8\%}} & \cellcolor[HTML]{EFEFEF}\textbf{-2.1\%} & \cellcolor[HTML]{EFEFEF}\textbf{-4.0\%} & \cellcolor[HTML]{EFEFEF}\textbf{-4.6\%} \\ \hline
\end{tabular}
\end{table*}

\begin{table*}[]
\caption{RD performance comparison between VCWF-M and GeS-TMv7.0-rc2 under C1 and C2 condition}
\setlength{\tabcolsep}{6pt}  
\renewcommand{\arraystretch}{0.97}
\centering
\label{table5}
\begin{tabular}{cllllllllllll}
\hline
                        & \multicolumn{12}{c}{End-to-End BD AttrRate (\%)}                                                                                                                                                                                                                                                                                                                                                                                                                                                                     \\
                        & \multicolumn{3}{c}{Octree-raht-intra-C1}                                                                         & \multicolumn{3}{c}{Octree-raht-inter-C1}                                                                         & \multicolumn{3}{c}{Octree-raht-intra-C2}                                                                                                         & \multicolumn{3}{c}{Octree-raht-inter-C2}                                                                                    \\
\multirow{-3}{*}{Class} & \multicolumn{1}{c}{Luma} & \multicolumn{1}{c}{Cb} & \multicolumn{1}{c}{Cr}                                       & \multicolumn{1}{c}{Luma} & \multicolumn{1}{c}{Cb} & \multicolumn{1}{c}{Cr}                                       & \multicolumn{1}{c}{Luma}                & \multicolumn{1}{c}{Cb}                  & \multicolumn{1}{c}{Cr}                                       & \multicolumn{1}{c}{Luma}                & \multicolumn{1}{c}{Cb}                  & \multicolumn{1}{c}{Cr}                  \\ \hline
\rowcolor[HTML]{EFEFEF} 
loot                    & -4.2\%                   & -3.1\%                 & \multicolumn{1}{l|}{\cellcolor[HTML]{EFEFEF}-4.8\%}          & -5.4\%                   & -8.6\%                 & \multicolumn{1}{l|}{\cellcolor[HTML]{EFEFEF}-7.9\%}          & \cellcolor[HTML]{EFEFEF}-2.5\%          & \cellcolor[HTML]{EFEFEF}-3.2\%          & \multicolumn{1}{l|}{\cellcolor[HTML]{EFEFEF}-5.4\%}          & \cellcolor[HTML]{EFEFEF}-1.0\%          & \cellcolor[HTML]{EFEFEF}-4.8\%          & \cellcolor[HTML]{EFEFEF}-4.8\%          \\
redandblack             & -5.0\%                   & -3.7\%                 & \multicolumn{1}{l|}{-5.0\%}                                  & -5.3\%                   & -3.9\%                 & \multicolumn{1}{l|}{-6.5\%}                                  & -3.6\%                                  & -3.6\%                                  & \multicolumn{1}{l|}{-1.6\%}                                  & -2.8\%                                  & -7.4\%                                  & -1.4\%                                  \\
\rowcolor[HTML]{EFEFEF} 
soldier                 & -4.4\%                   & -2.3\%                 & \multicolumn{1}{l|}{\cellcolor[HTML]{EFEFEF}-2.6\%}          & -3.2\%                   & -9.2\%                 & \multicolumn{1}{l|}{\cellcolor[HTML]{EFEFEF}-9.6\%}          & \cellcolor[HTML]{EFEFEF}-1.8\%          & \cellcolor[HTML]{EFEFEF}-2.6\%          & \multicolumn{1}{l|}{\cellcolor[HTML]{EFEFEF}-3.6\%}          & \cellcolor[HTML]{EFEFEF}0.8\%           & \cellcolor[HTML]{EFEFEF}-9.3\%          & \cellcolor[HTML]{EFEFEF}-9.8\%          \\
queen                   & \textbf{-9.7\%}          & \textbf{-12.2\%}       & \multicolumn{1}{l|}{\textbf{-12.2\%}}                        & \textbf{-13.5\%}         & \textbf{-20.3\%}       & \multicolumn{1}{l|}{\textbf{-19.5\%}}                        & \textbf{-5.3\%}                         & \textbf{-6.5\%}                         & \multicolumn{1}{l|}{\textbf{-5.1\%}}                         & \textbf{-8.3\%}                         & \textbf{-11.0\%}                        & \textbf{-8.7\%}                         \\
\rowcolor[HTML]{EFEFEF} 
longdress               & -3.2\%                   & -5.3\%                 & \multicolumn{1}{l|}{\cellcolor[HTML]{EFEFEF}-5.4\%}          & -3.2\%                   & -4.9\%                 & \multicolumn{1}{l|}{\cellcolor[HTML]{EFEFEF}-5.7\%}          & \cellcolor[HTML]{EFEFEF}-1.1\%          & \cellcolor[HTML]{EFEFEF}-3.5\%          & \multicolumn{1}{l|}{\cellcolor[HTML]{EFEFEF}-2.3\%}          & \cellcolor[HTML]{EFEFEF}-0.8\%          & \cellcolor[HTML]{EFEFEF}-3.7\%          & \cellcolor[HTML]{EFEFEF}-2.8\%          \\
basketball\_player      & -5.2\%                   & -1.2\%                 & \multicolumn{1}{l|}{-3.1\%}                                  & -6.0\%                   & -2.3\%                 & \multicolumn{1}{l|}{-4.5\%}                                  & -3.6\%                                  & -0.4\%                                  & \multicolumn{1}{l|}{-3.0\%}                                  & -4.5\%                                  & 0.8\%                                   & -7.8\%                                  \\
\rowcolor[HTML]{EFEFEF} 
dancer\_player          & -5.4\%                   & -1.6\%                 & \multicolumn{1}{l|}{\cellcolor[HTML]{EFEFEF}-3.0\%}          & -5.9\%                   & -2.1\%                 & \multicolumn{1}{l|}{\cellcolor[HTML]{EFEFEF}-2.7\%}          & \cellcolor[HTML]{EFEFEF}-2.9\%          & \cellcolor[HTML]{EFEFEF}-0.9\%          & \multicolumn{1}{l|}{\cellcolor[HTML]{EFEFEF}-2.2\%}          & \cellcolor[HTML]{EFEFEF}-3.5\%          & \cellcolor[HTML]{EFEFEF}-1.1\%          & \cellcolor[HTML]{EFEFEF}-3.9\%          \\ \hline
Cat2-A average          & -5.8\%                   & -5.3\%                 & \multicolumn{1}{l|}{-6.1\%}                                  & -6.9\%                   & -10.5\%                & \multicolumn{1}{l|}{-10.9\%}                                 & -3.3\%                                  & -4.0\%                                  & \multicolumn{1}{l|}{-3.9\%}                                  & -2.9\%                                  & -8.1\%                                  & -6.2\%                                  \\
\rowcolor[HTML]{EFEFEF} 
Cat2-B average          & -3.2\%                   & -5.3\%                 & \multicolumn{1}{l|}{\cellcolor[HTML]{EFEFEF}-5.4\%}          & -3.2\%                   & -4.9\%                 & \multicolumn{1}{l|}{\cellcolor[HTML]{EFEFEF}-5.7\%}          & \cellcolor[HTML]{EFEFEF}-1.1\%          & \cellcolor[HTML]{EFEFEF}-3.5\%          & \multicolumn{1}{l|}{\cellcolor[HTML]{EFEFEF}-2.3\%}          & \cellcolor[HTML]{EFEFEF}-0.8\%          & \cellcolor[HTML]{EFEFEF}-3.7\%          & \cellcolor[HTML]{EFEFEF}-2.8\%          \\
Cat2-C average          & -5.3\%                   & -1.4\%                 & \multicolumn{1}{l|}{-3.1\%}                                  & -5.9\%                   & -2.2\%                 & \multicolumn{1}{l|}{-3.6\%}                                  & -3.2\%                                  & -0.7\%                                  & \multicolumn{1}{l|}{-2.6\%}                                  & -4.0\%                                  & -0.2\%                                  & -5.8\%                                  \\
\rowcolor[HTML]{EFEFEF} 
Overall average         & \textbf{-5.3\%}          & \textbf{-4.2\%}        & \multicolumn{1}{l|}{\cellcolor[HTML]{EFEFEF}\textbf{-5.1\%}} & \textbf{-6.1\%}          & \textbf{-7.3\%}        & \multicolumn{1}{l|}{\cellcolor[HTML]{EFEFEF}\textbf{-8.0\%}} & \cellcolor[HTML]{EFEFEF}\textbf{-3.0\%} & \cellcolor[HTML]{EFEFEF}\textbf{-3.0\%} & \multicolumn{1}{l|}{\cellcolor[HTML]{EFEFEF}\textbf{-3.3\%}} & \cellcolor[HTML]{EFEFEF}\textbf{-3.0\%} & \cellcolor[HTML]{EFEFEF}\textbf{-5.2\%} & \cellcolor[HTML]{EFEFEF}\textbf{-5.6\%} \\ \hline
\end{tabular}
\end{table*}

\subsection{Objective Quality Evaluation}
To evaluate the RD performance of the proposed method quantitively, we used bits per output point (BPOP) as the coding bitrate and peak signal-to-noise ratio (PSNR) as the quality of the reconstructed attribute \cite{ref44}. For the same reconstruction quality, the lower the BPOP, the better the performance, or rather, for the same coding bitrate, the larger the PSNR, the better the performance. To verify the RD performance comprehensively, six test bitrates, namely r01, r02, r03, r04, r05, and r06, are recommended by the CTC, and Bjøntegaard Delta Rate (BD-rate) \cite{ref44} is adopted as a quantitative measure for the overall RD performance. The BD-rate is a percentage, showing how much bitrate one encoding method increases over another while achieving the same PSNR. Therefore, a negative BD-rate indicates better RD performance. As the proposed Wiener filter is for attribute, we only consider the bitrate and reconstruction quality of attribute, thereby, the BD-rate is also referred to as “End-to-End BD AttrRate (\%)” in the following. 

Since the proposed method includes four parts, i.e., M-KNN, BWF, CIWF, and VCWF, to compare the performance, the BWF with M-KNN is named BWF-M. Similarly, CIWF-M and VCWF-M indicate CIWF with M-KNN and VCWF with M-KNN, respectively. Tables \ref{table3}, \ref{table4}, and \ref{table5} compares the End-to-End BD AttrRates (\%) of BWF-M, CIWF-M, VCWF-M with GeS-TMv7.0-rc2 in terms of both intra and inter prediction under C1 and C2 test condition, respectively. From these tables, we can see that, the RD performance for inter prediction is better than that for intra prediction. This is because quality enhancement can not only improve the quality of a specific frame but also eliminate accumulated coding distortion for successive frames to some extent. It can also be seen that the performance of the C2 test condition is inferior to that of the C1 test condition. This is primarily because the C2 condition involves a smaller bitstream, which makes the impact of bits of the filter coefficients more significant compared to the C1 condition. From Table \ref{table4}, we can see that CIWF-M is efficient for both intra and inter prediction and for all the color components. From Table \ref{table5}, we can see that the RD performance of Luma component can be significantly improved by VCWF-M, while the performance of the Chroma components is slightly degraded. This is because the coding bits of the color components are interleaved and cannot be divided from the attribute bitstream, and thus we can only use the total coding bits to calculate the BD-rates for all the color components. Accordingly, when more filter coefficients for Luma component are encoded into the bitstream, the RD performance of Chroma components will also be affected although the distortion of Chroma components does not change. We can see that the performance of “\emph{queen}" is particularly well.  The reason is that “\emph{queen}" is a computer-generated point cloud that has a stronger inter-frame similarity compared to device captured point clouds which contain temporal noises inevitably.

\subsection{Subjective Quality Evaluation}
\begin{figure}[!t]
\includegraphics[width=0.48\textwidth]{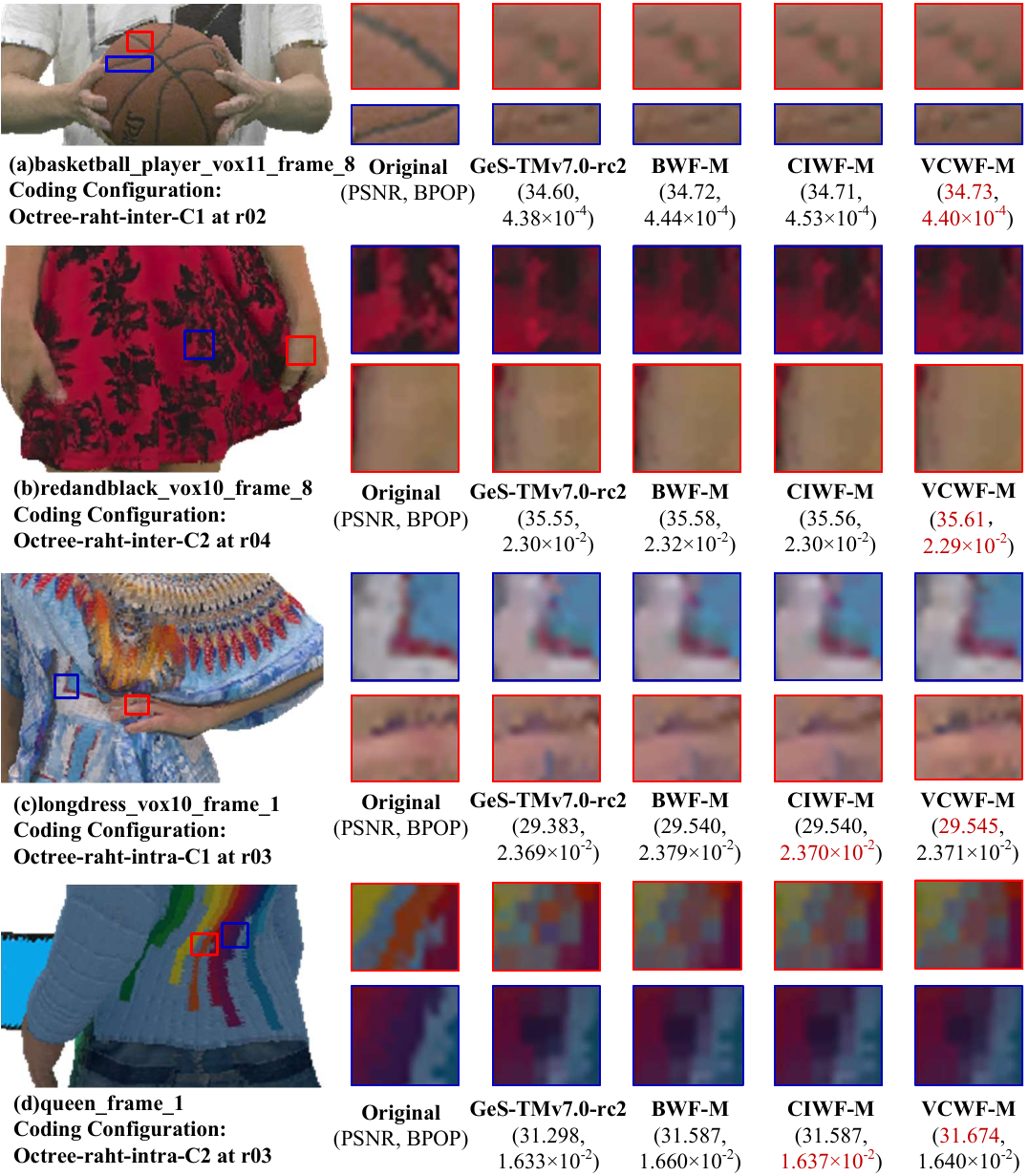}
\caption{Subjective quality comparison. Left: Original point cloud. Right: Reconstructed point cloud using GeS-TMv7.0-rc2, BWF-M, CIWF-M, and VCWF-M. The weighted average (7:1:1) of PSNRs (dB) of Luma, Chroma Cb, and Chroma Cr and BPOP of each frame are given below each sub-figure.}
\label{fig13}
\end{figure}

To demonstrate the effectiveness of the proposed method in terms of subjective quality, we randomly selected several snapshots from different dynamic point clouds, including “\emph{basketball\_player\_vox11\_frame\_8.ply}”, “\emph{redandblack\_vox10\_frame\_8.ply}”, “\emph{longdress\_vox10\_frame\_1.ply}” and “\emph{queen\_frame\_1.ply}”. We then compared the reconstruction quality of these snapshots under different configurations, both before and after applying the proposed Wiener filters, as shown in Fig. \ref{fig13}. 

We can observe that the filtered point clouds are smoother than the unfiltered point clouds in smooth regions, while in textured regions, the filtered point clouds are finer, indicating the effectiveness of the proposed method. We can also see that BWF-M and CIWF-M show similar enhancement effects, indicating that the CIWF-M can preserve the performance while consuming fewer bits. Among all the methods, VCWF-M provides the best quality with the lowest bit consumption. For inter-frame prediction, the single-frame BPOP is even lower than the anchor (i.e., GeS-TMv7.0-rc2). This is because the quality enhanced reference frame can lead to better prediction accuracy for dynamic point clouds, which in turn reduces the bits required to encode the residuals.

\begin{table*}[]
\caption{Complexity ratio compared with GeS-TMv7.0-rc2}
\centering
\setlength{\tabcolsep}{6pt}  
\label{table6}
\begin{tabular}{ccccccccc}
\hline
                         & \multicolumn{8}{c}{Complexity Ratio (\%)}                                                                                                                                                                                                                                             \\
                         & \multicolumn{2}{c}{Octree-raht-inter-C1}                                     & \multicolumn{2}{c}{Octree-raht-inter-C2}                                     & \multicolumn{2}{c}{Octree-raht-intra-C1}                                     & \multicolumn{2}{c}{Octree-raht-intra-C2} \\
\multirow{-3}{*}{Method} & $CR_{encoder}$ & $CR_{decoder}$                                              & $CR_{encoder}$ & $CR_{decoder}$                                              & $CR_{encoder}$ & $CR_{decoder}$                                              & $CR_{encoder}$      & $CR_{decoder}$     \\ \hline
\rowcolor[HTML]{EFEFEF} 
BWF + M-KNN              & 111\%          & \multicolumn{1}{c|}{\cellcolor[HTML]{EFEFEF}\textbf{108\%}} & 105\%          & \multicolumn{1}{c|}{\cellcolor[HTML]{EFEFEF}\textbf{107\%}} & 125\%          & \multicolumn{1}{c|}{\cellcolor[HTML]{EFEFEF}\textbf{113\%}} & 107\%               & \textbf{110\%}     \\
\rowcolor[HTML]{FFFFFF} 
BWF + KNN                & 184\%          & \multicolumn{1}{c|}{\cellcolor[HTML]{FFFFFF}204\%}          & 134\%          & \multicolumn{1}{c|}{\cellcolor[HTML]{FFFFFF}178\%}          & 273\%          & \multicolumn{1}{c|}{\cellcolor[HTML]{FFFFFF}247\%}          & 141\%               & 208\%              \\
\rowcolor[HTML]{EFEFEF} 
CIWF + M-KNN             & \textbf{109\%} & \multicolumn{1}{c|}{\cellcolor[HTML]{EFEFEF}109\%}          & \textbf{102\%} & \multicolumn{1}{c|}{\cellcolor[HTML]{EFEFEF}110\%}          & \textbf{119\%} & \multicolumn{1}{c|}{\cellcolor[HTML]{EFEFEF}117\%}          & \textbf{103\%}      & 117\%              \\
\rowcolor[HTML]{FFFFFF} 
CIWF + KNN               & 182\%          & \multicolumn{1}{c|}{\cellcolor[HTML]{FFFFFF}205\%}          & 130\%          & \multicolumn{1}{c|}{\cellcolor[HTML]{FFFFFF}193\%}          & 268\%          & \multicolumn{1}{c|}{\cellcolor[HTML]{FFFFFF}254\%}          & 136\%               & 234\%              \\
\rowcolor[HTML]{EFEFEF} 
VCWF + M-KNN             & 110\%          & \multicolumn{1}{c|}{\cellcolor[HTML]{EFEFEF}114\%}          & 102\%          & \multicolumn{1}{c|}{\cellcolor[HTML]{EFEFEF}115\%}          & 124\%          & \multicolumn{1}{c|}{\cellcolor[HTML]{EFEFEF}125\%}          & 104\%               & 124\%              \\
\rowcolor[HTML]{FFFFFF} 
VCWF + KNN               & 187\%          & \multicolumn{1}{c|}{\cellcolor[HTML]{FFFFFF}221\%}          & 131\%          & \multicolumn{1}{c|}{\cellcolor[HTML]{FFFFFF}214\%}          & 280\%          & \multicolumn{1}{c|}{\cellcolor[HTML]{FFFFFF}273\%}          & 140\%               & 262\%              \\ \hline
\end{tabular}
\end{table*}

\subsection{Complexity Comparison}
\label{sub-sec:Complexity Comparison}
Table \ref{table6} compares the complexity ratio of the proposed method with GeS-TMv7.0-rc2 at encoder and decoder. The average encoding and decoding complexity ratios can be calculated as
\begin{equation}
\label{deqn_ex21}
CR_{encoder(resp.\ decoder)} = \frac{T_{encoder(resp.\ decoder)}^{proposed}}{T_{encoder(resp.\ decoder)}^{anchor}} \times 100 \%,
\end{equation}
where $T_{encoder(resp.\ decoder)}^{proposed}$ and $T_{encoder(resp.\ decoder)}^{anchor}$ represent the average encoding (resp. decoding) time of the proposed method and the anchor (i.e., GeS-TMv7.0-rc2), respectively. Besides, we also compare the effectiveness of M-KNN with KNN in this table. 

We can see that Wiener filter bring additional complexity to both encoder and decoder. Compared with KNN, M-KNN significantly reduces the time required for nearest neighbor search. Among the proposed methods, CIWF-M requires the lowest encoding time complexity, because it can not only avoid a lot of Wiener filter coefficients calculation in BWF-M but also does not need to carry out the variance-based classification for Luma in VCWF-M. 

The encoding complexity increment of our proposed method under the C1 condition is higher than the C2 condition. This is because the time complexity of the matrix multiplication involved in the filtering process is related to the number of points. Specifically, the more points there are, the greater the complexity is. Since the reconstructed point clouds under the C1 condition (geometry is encoded losslessly) contain more points, the increase of time complexity is more pronounced. Additionally, the complexity increment of inter-frame coding is smaller than that of intra-frame coding. This is due to the fact that intra-frame coding has a lower encoding time, and the complexity introduced by including the Wiener filter is more significant in intra-frame coding. 
\section{Conclusion}
\label{sec:Conclusion}
We proposed a high efficiency Wiener filter method to enhance the reconstruction quality as well as improve the RD performance for MPEG G-PCC. It includes four parts, i.e., M-KNN for fast neighbor search, BWF for basic implementation of Wiener filter, CIWF which saves filter coefficients by inheriting them from the first frame, VCWF which is specifically designed based on point classification for Luma component. Experimental results demonstrate that all the four parts are effective, and by integrating them together (i.e., VCWF-M), significant RD performance gain in terms of both objective and subjective metrics can be achieved by comparing to the latest test platform of G-PCC with affordable complexity cost. In the future, we will refine the proposed method to further improve the coding efficiency and reduce time complexity. 
{\appendix[Distortion Analysis of Inter Prediction]
For efficient compression of dynamic point clouds, inter prediction is utilized to exploit the temporal correlations between successive frames. It aims to reduce redundancy by leveraging similarities between temporal adjacent frames. Therefore, the distortion of the previous frame will directly affect the reconstruction quality of the subsequent frames. For two successive frames $frame_{ref}$  and $frame_{cur}$ whose attributes are denoted as $\bm{A} _{ref}$  and $\bm{A} _{cur}$, respectively. Their reconstructed attributes are denoted as $\bm{\hat{A} } _{ref}$  and $\bm{\hat{A} } _{cur}$, then the coding error of $frame_{ref}$ can be denoted as
\begin{equation}
\label{deqn_ex14}
\bm{D} _{ref} = \bm{A} _{ref} - \bm{\hat{A} } _{ref}.
\end{equation}

Due to the inter prediction, the reconstructed attribute $\bm{\hat{A} } _{cur}$ can be expressed as
\begin{equation}
\begin{aligned}
\label{deqn_ex15}
\bm{\hat{A} } _{cur} &= \bm{\hat{A} }_{ref} + \Delta \bm{A}  \\
                     &= \bm{A} _{ref} - \bm{D} _{ref} + \Delta \bm{A},
\end{aligned}
\end{equation}
where $\Delta \bm{A}$ represents the prediction error of $frame_{cur}$ relative to $frame_{ref}$. Then the attribute coding error $\bm{D} _{cur}$ can be denoted as
\begin{equation}
\begin{aligned}
\label{deqn_ex16}
\bm{D}_{cur} &= \bm{A}_{cur} - \hat{\bm{A}} _{cur} \\
             &= \bm{A}_{cur} - \bm{A}_{ref} - \Delta \bm{A} + \bm{D}_{ref}.
\end{aligned}
\end{equation}

We can see that $\bm{D} _{cur}$ not only depends on the attributes of $frame_{ref}$ and $frame_{cur}$ but also depends on $\bm{D} _{ref}$. If $\bm{D} _{ref}$ decreases, $\bm{D} _{cur}$ will also decrease. Therefore, the quality improvement in one frame can also benefit the subsequent frames.} 
 
\begin{IEEEbiography}[{\includegraphics[width=1in,height=1.25in,clip,keepaspectratio]{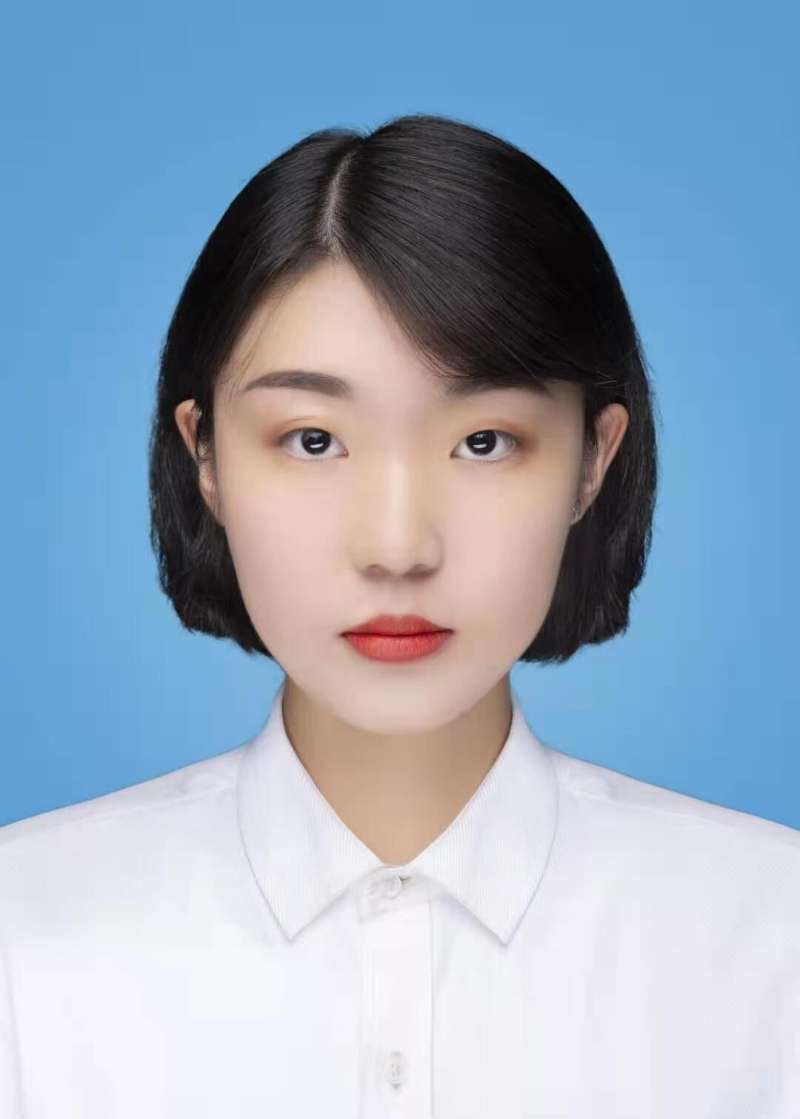}}]{Yuxuan Wei}
received the B.E. degree in automation with the Department of Control Science and Engineering from Shandong University, Ji’nan, China, in 2023. She is now pursuing the M.E. degree in Control Science and Engineering from Shandong University, Ji’nan, China. Her research interests include 3D point cloud compression and post processing.
\end{IEEEbiography}

\begin{IEEEbiography}[{\includegraphics[width=1in,height=1.25in,clip,keepaspectratio]{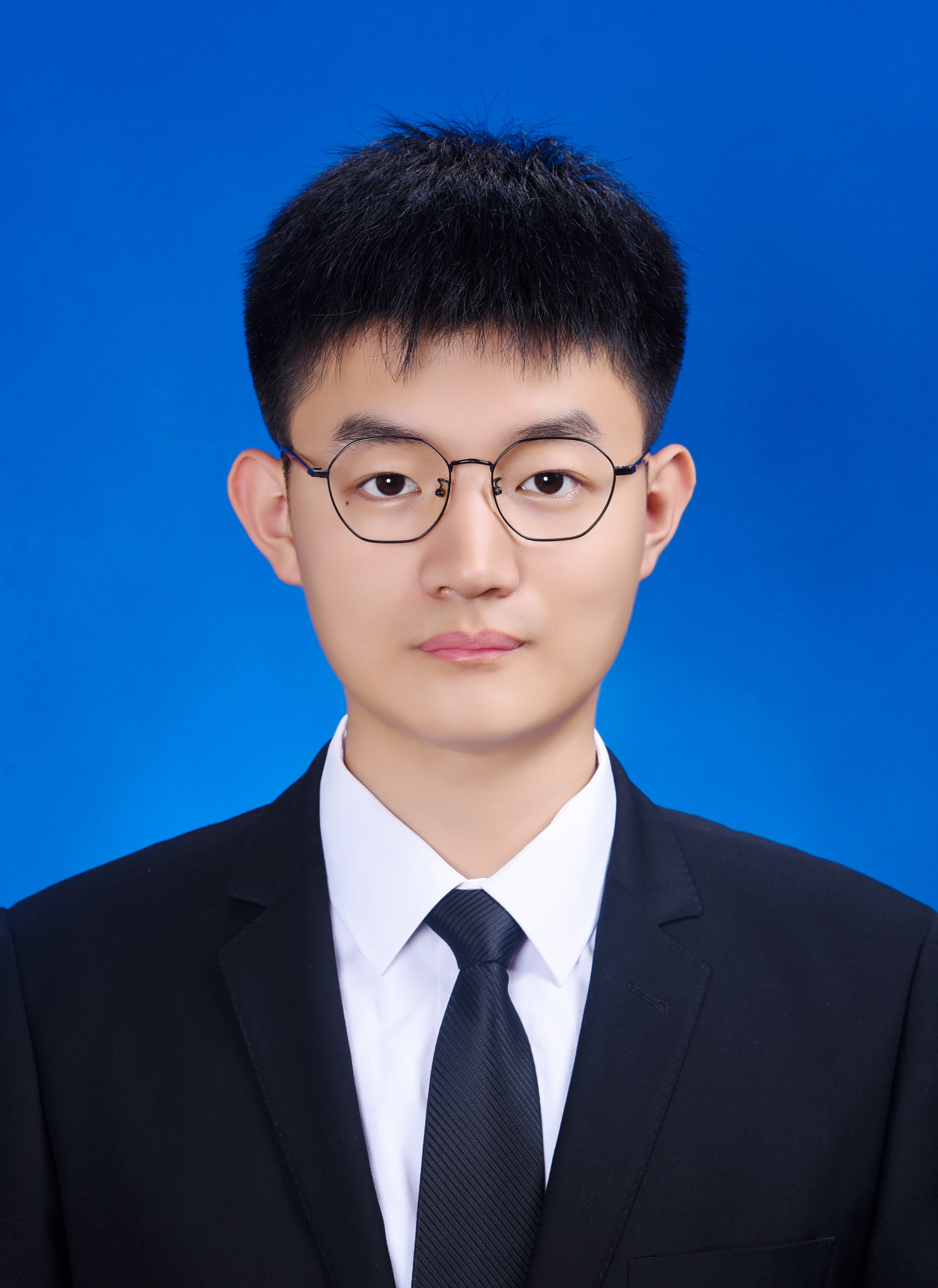}}]{Zehan Wang}
received the B.E. degree in electronic information engineering from the School of Automation and Information Engineering, Xi'an University of Technology, Xi'an, China, in 2023. He is currently pursuing the M.E. degree in the Department of Control Science and Engineering, Shandong University, Ji’nan, China. He is actively participating in the development of point cloud compression standards. His research interests include 3D point cloud compression and processing.
\end{IEEEbiography}

\begin{IEEEbiography}[{\includegraphics[width=1in,height=1.25in,clip,keepaspectratio]{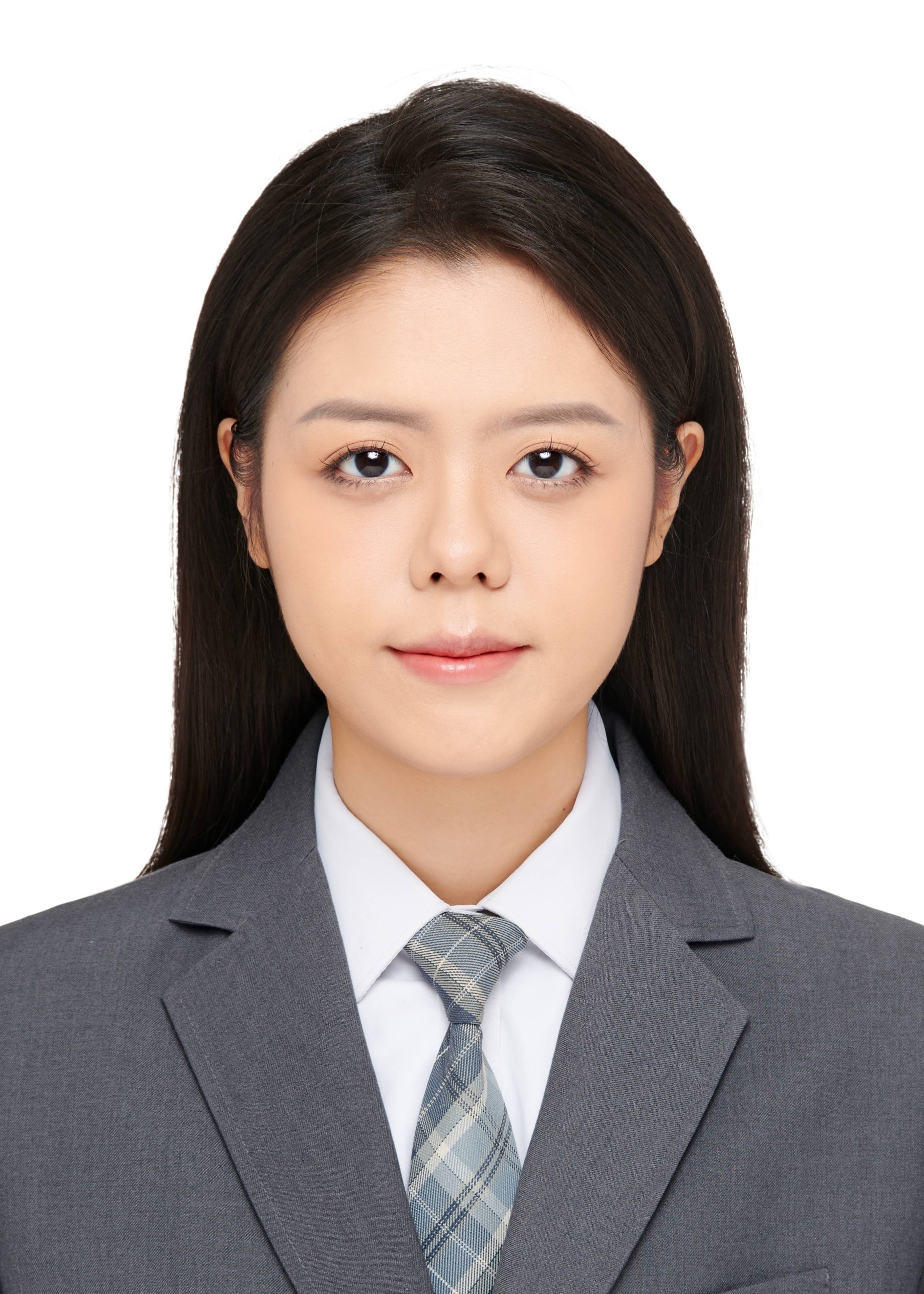}}]{Tian Guo}
received the B.E. degree from the School of Information and Control Engineering, China University of Mining and Technology, Jiangsu, China, in 2021. She is currently pursuing the Ph.D. degree in Control Science and Engineering with Shandong University, Shandong. Her research interests include point clouds compression and processing.
\end{IEEEbiography}

\begin{IEEEbiography}[{\includegraphics[width=1in,height=1.25in,clip,keepaspectratio]{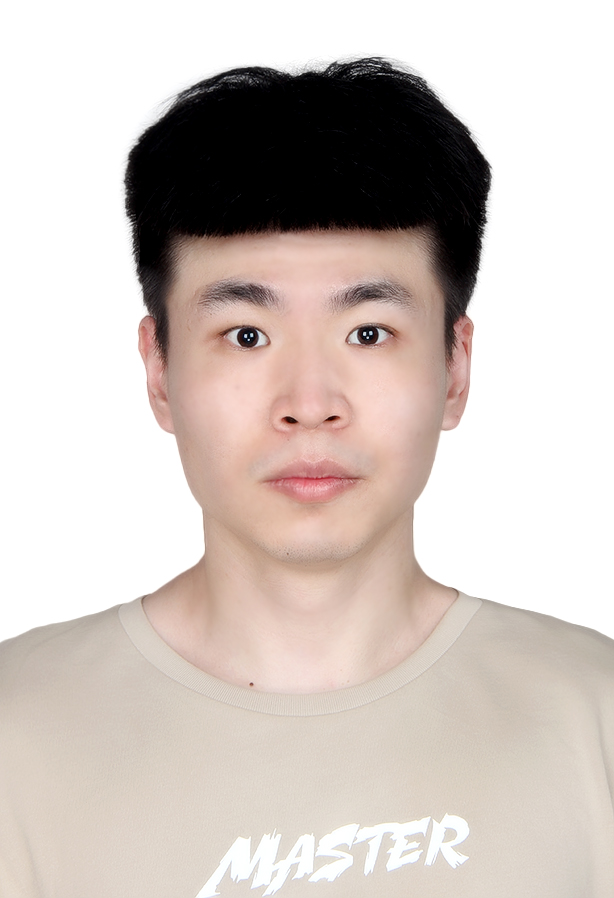}}]{Hao Liu}
received the B.E. degree in telecommunication engineering from Shandong Agricultural University, Taian, China, in 2017. He received the Ph.D. degree in information science and engineering from Shandong University, Qingdao, China, in 2022. He has been a lecturer with School of Computer Science and Control Engineering, Yantai University, since July 2022. His research interests include 3D point clouds compression and processing.
\end{IEEEbiography}

\begin{IEEEbiography}[{\includegraphics[width=1in,height=1.25in,clip,keepaspectratio]{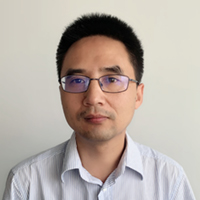}}]{Liquan Shen}
received the B.E. degree in automation control from Henan Polytechnic University, Henan, China, in 2001, and the M.E. and Ph.D. degrees in communication and information systems from Shanghai University, Shanghai, China, in 2005 and 2008,respectively.Since2008,hehasbeenaFaculty Member with the School of Communication and Information Engineering, Shanghai University, where he is currently a Professor. From 2013 to 2014, he was a Visiting Professor with the Department of Electrical and Computer Engineering, University of Florida, Gainesville, FL, USA. He has authored or coauthored more than 100 refereed technical papers in international journals and conferences in the fields of video coding and image processing. He holds ten patents in the areas of image or video coding and communications. His research interests include versatile video coding (VVC), perceptual coding, video codec optimization, 3DTV, and video quality assessment.
\end{IEEEbiography}

\begin{IEEEbiography}[{\includegraphics[width=1in,height=1.25in,clip,keepaspectratio]{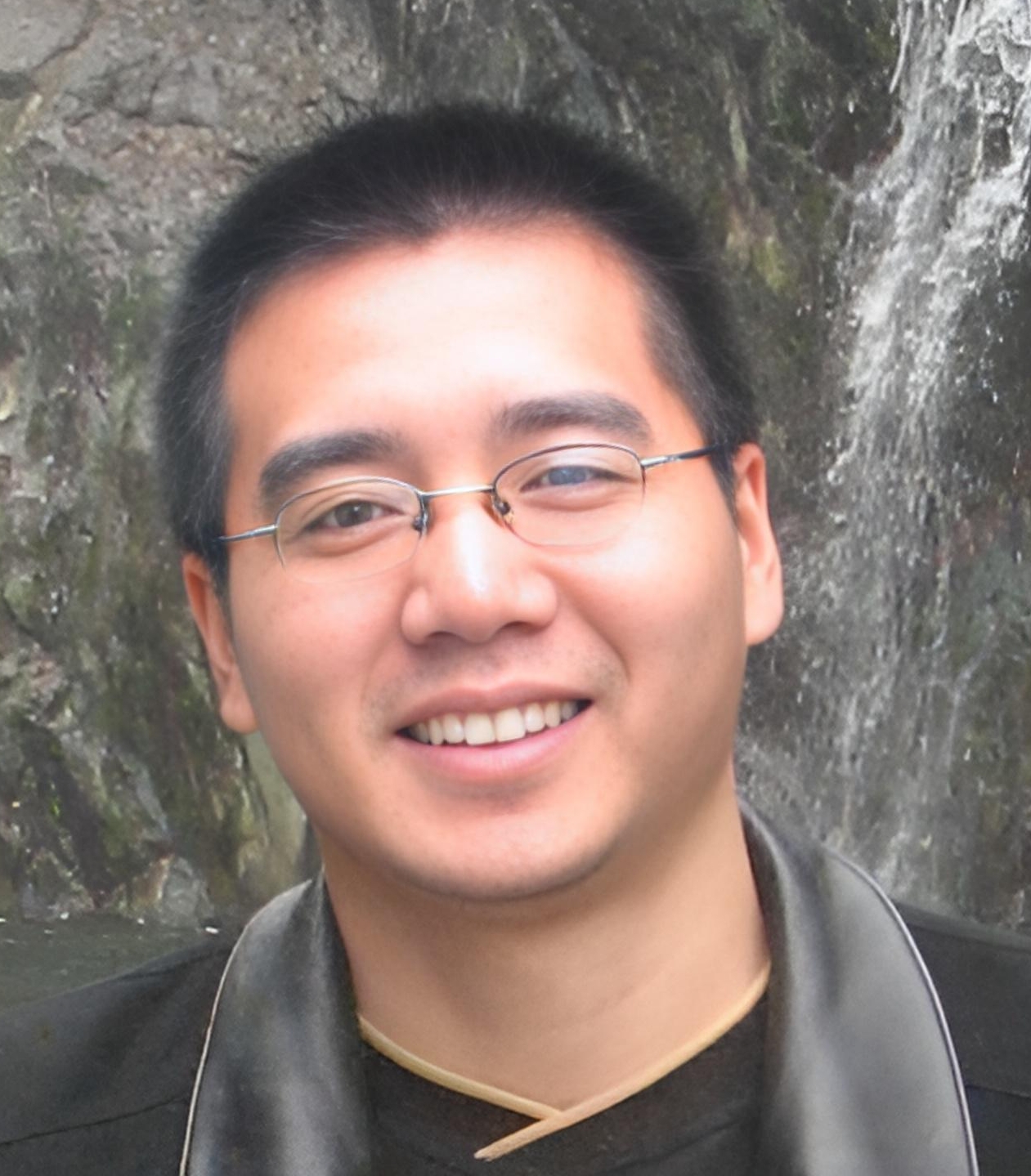}}]{Hui Yuan}
(Senior Member, IEEE) received the B.E. and Ph.D. degrees in telecommunication engineering from Xidian University, Xi’an, China, in 2006 and 2011, respectively. In April 2011, he joined Shandong University, Ji’nan, China, as a Lecturer (April 2011–December 2014), an Associate Professor (January 2015-August 2016), and a Professor (September 2016). From January 2013 to December 2014, and from November 2017 to February 2018, he worked as a Postdoctoral Fellow (Granted by the Hong Kong Scholar Project) and a Research Fellow, respectively, with the Department of Computer Science, City University of Hong Kong. From November 2020 to November 2021, he worked as a Marie Curie Fellow (Granted by the Marie Skłodowska-Curie Actions Individual Fellowship under Horizon2020 Europe) with the School of Engineering and Sustainable Development, De Montfort University, Leicester, U.K. From October 2021 to November 2021, he also worked as a visiting researcher (secondment of the Marie Skłodowska-Curie Individual Fellowships) with the Computer Vision and Graphics group, Fraunhofer Heinrich-Hertz-Institut (HHI), Germany. His current research interests include 3D visual media coding, processing, and communication.

Prof. Yuan serves as an Associate Editor for IEEE Transactions on Image Processing (since 2025), an Associate Editor for IEEE Transactions on Consumer Electronics (since 2024), an Associate Editor for IET Image Processing (since 2023) and an Area Chair for IEEE ICME (since 2020).
\end{IEEEbiography}


\begin{thebibliography}{1}
\bibliographystyle{IEEEtran}

\bibitem{ref1}
Z. Wang, Y. Wei, H. Yuan, W. Zhang and P. Li, ``Rate-Distortion Optimized Skip Coding of Region Adaptive Hierarchical Transform Coefficients for MPEG G-PCC,'' in \textit{IEEE Transactions on Circuits and Systems for Video Technology}, doi: 10.1109/TCSVT.2024.3487543.

\bibitem{ref2}
J. Zhang, J. Zhang, W. Ma, D. Ding and Z. Ma, ``Content-Aware Rate Control for Geometry-Based Point Cloud Compression,'' in \textit{IEEE Transactions on Circuits and Systems for Video Technology}, vol. 34, no. 10, pp. 9550-9561, Oct. 2024.

\bibitem{ref3}
M. Xu et al., ``A Full Dive Into Realizing the Edge-Enabled Metaverse: Visions, Enabling Technologies, and Challenges,'' in \textit{IEEE Communications Surveys \& Tutorials}, vol. 25, no. 1, pp. 656-700, 2023. 

\bibitem{ref4}
H. Liu, H. Yuan, Q. Liu, J. Hou and J. Liu, ``A Comprehensive Study and Comparison of Core Technologies for MPEG 3-D Point Cloud Compression,'' in \textit{IEEE Transactions on Broadcasting}, vol. 66, no. 3, pp. 701-717, Sept. 2020.

\bibitem{ref5}
S. Schwarz et al., ``Emerging MPEG Standards for Point Cloud Compression,'' in \textit{IEEE Journal on Emerging and Selected Topics in Circuits and Systems}, vol. 9, no. 1, pp. 133-148, March 2019. 

\bibitem{ref6}
MPEG 3D Graphics Coding and Haptics Coding, ``G-PCC Codec Description,'' in \textit{ISO/IEC JTC1/SC29/WG11 MPEG output document w19332}, Alpbach, Apr. 2020.

\bibitem{ref7}
MPEG 3D Graphics Coding and Haptics Coding, ``V-PCC Codec Description,'' in \textit{ISO/IEC JTC1/SC29/WG11 MPEG output document w19331}, Alpbach, Apr. 2020.

\bibitem{ref8}
MPEG 3D Graphics Coding and Haptics Coding, ``EE 13.60 on dynamic solid coding with G-PCC,'' in \textit{ISO/IEC JTC1/SC29/WG07 MPEG output document N00528}, Online, Jan. 2023.

\bibitem{ref9}
S. Lasserre, ``Improving TriSoup summary, results and perspective,'' in \textit{ISO/IEC JTC1/SC29/WG07 MPEG input document m59288}, Online, Apr.2022.

\bibitem{ref10}
Z. Wang, S. Wan and L. Wei, ``Local Geometry-Based Intra Prediction for Octree-Structured Geometry Coding of Point Clouds,'' in \textit{IEEE Transactions on Circuits and Systems for Video Technology}, vol. 33, no. 2, pp. 886-896, Feb. 2023.

\bibitem{ref11}
R. L. de Queiroz and P. A. Chou, ``Compression of 3D Point Clouds Using a Region-Adaptive Hierarchical Transform,'' in \textit{IEEE Transactions on Image Processing}, vol. 25, no. 8, pp. 3947-3956, Aug. 2016. 

\bibitem{ref12}
W. Zhang, F. Yang, Y. Xu and M. Preda, ``Standardization Status of MPEG Geometry-Based Point Cloud Compression (G-PCC) Edition 2,'' in \textit{2024 Picture Coding Symposium}, Taichung, Taiwan, 2024, pp. 1-5.

\bibitem{ref13}
A. L. Souto, R. L. De Queiroz and C. Dorea, ``Motion-Compensated Predictive RAHT for Dynamic Point Clouds,'' in \textit{IEEE Transactions on Image Processing}, vol. 32, pp. 2428-2437, 2023.

\bibitem{ref14}
MPEG 3D Graphics Coding and Haptics Coding, ``EE 13.2 on inter prediction,'' in \textit{ISO/IEC JTC1/SC29/WG07 MPEG output document N00943}, Sapporo, July. 2024.

\bibitem{ref15}
L. Wang, J. Sun, H. Yuan, R. Hamzaoui and X. Wang, ``Kalman filter-based prediction refinement and quality enhancement for geometry-based point cloud compression,'' in \textit{2021 International Conference on Visual Communications and Image Processing}, Munich, Germany, 2021, pp. 1-5.

\bibitem{ref16}
J. Xing, H. Yuan, C. Chen and W. Gao, ``Wiener Filter-based Color Attribute Quality Enhancement for Geometry-based Point Cloud Compression,'' in \textit{2022 Asia-Pacific Signal and Information Processing Association Annual Summit and Conference}, Chiang Mai, Thailand, 2022, pp. 1208-1212.

\bibitem{ref17}
J. Xing, H. Yuan, C. Chen, and T. Guo, ``Wiener filter-based point cloud adaptive denoising for video-based point cloud compression,'' in \textit{Proc. 1st Int. Workshop Adv. Point Cloud Compress., Process. Anal.}, New York, USA, Oct. 2022, pp. 21–25.

\bibitem{ref18}
W. -C. Lin et al., ``3D Point Cloud Denoising Based on Color Attribute,'' in \textit{2023 Asia Pacific Signal and Information Processing Association Annual Summit and Conference}, Taipei, Taiwan, 2023, pp. 1512-1516.

\bibitem{ref19}
K. Yamamoto, M. Onuki and Y. Tanaka, ``Deblurring of point cloud attributes in graph spectral domain,'' in \textit{2016 IEEE International Conference on Image Processing}, Phoenix, AZ, USA, 2016, pp. 1559-1563.

\bibitem{ref20}
F. Xue, F. Luisier and T. Blu, ``Multi-Wiener SURE-LET Deconvolution,'' in \textit{IEEE Transactions on Image Processing}, vol. 22, no. 5, pp. 1954-1968, May 2013. 

\bibitem{ref21}
C. Dinesh, G. Cheung and I. V. Bajić, ``3D Point Cloud Color Denoising Using Convex Graph-Signal Smoothness Priors,'' in \textit{2019 IEEE 21st International Workshop on Multimedia Signal Processing}, Kuala Lumpur, Malaysia, 2019, pp. 1-6.

\bibitem{ref22}
R. Watanabe, K. Nonaka, E. Pavez, T. Kobayashi and A. Ortega, ``Graph-Based Point Cloud Color Denoising with 3-Dimensional Patch-Based Similarity,'' in \textit{ICASSP 2023 - 2023 IEEE International Conference on Acoustics, Speech and Signal Processing}, Rhodes Island, Greece, 2023, pp. 1-5.

\bibitem{ref23}
M. A. Irfan and E. Magli, ``3D Point Cloud Denoising Using a Joint Geometry and Color k-NN Graph,'' in \textit{2020 28th European Signal Processing Conference}, Amsterdam, Netherlands, 2021, pp. 585-589.

\bibitem{ref24}
M. A. Irfan and E. Magli, ``Point Cloud Denoising using Joint Geometry/Color Graph Wavelets,'' in \textit{2020 IEEE Workshop on Signal Processing Systems}, Coimbra, Portugal, 2020, pp. 1-6.

\bibitem{ref25}
W. Tao, G. Jiang, M. Yu, Y. Zhang, Z. Jiang and Y. -S. Ho, ``Multi-View Projection Based Joint Geometry and Color Hole Repairing Method for G-PCC Trisoup Encoded Color Point Cloud,'' in \textit{IEEE Transactions on Emerging Topics in Computational Intelligence}, vol. 8, no. 1, pp. 892-902, Feb. 2024.

\bibitem{ref26}
X. Sheng, L. Li, D. Liu and Z. Xiong, ``Attribute Artifacts Removal for Geometry-Based Point Cloud Compression,'' in \textit{IEEE Transactions on Image Processing}, vol. 31, pp. 3399-3413, 2022.

\bibitem{ref27}
J. Xing, H. Yuan, R. Hamzaoui, H. Liu and J. Hou, ``GQE-Net: A Graph-Based Quality Enhancement Network for Point Cloud Color Attribute,'' in \textit{IEEE Transactions on Image Processing}, vol. 32, pp. 6303-6317, 2023. 

\bibitem{ref28}
J. Xing, H. Yuan, W. Zhang, T. Guo and C. Chen, ``A Small-Scale Image U-Net-Based Color Quality Enhancement for Dense Point Cloud,'' in \textit{IEEE Transactions on Consumer Electronics}, vol. 70, no. 1, pp. 669-683, Feb. 2024. 

\bibitem{ref29}
Y. Schoenenberger, J. Paratte and P. Vandergheynst, ``Graph-based denoising for time-varying point clouds,'' in \textit{2015 3DTV-Conference: The True Vision - Capture, Transmission and Display of 3D Video}, Lisbon, Portugal, 2015, pp. 1-4.

\bibitem{ref30}
W. Hu, Q. Hu, Z. Wang and X. Gao, ``Dynamic Point Cloud Denoising via Manifold-to-Manifold Distance,'' in \textit{IEEE Transactions on Image Processing}, vol. 30, pp. 6168-6183, 2021.

\bibitem{ref31}
X. Yan, J. Yang, X. Zhu, Y. Liang and H. Huang, ``Denoising Framework Based on Multiframe Continuous Point Clouds for Autonomous Driving LiDAR in Snowy Weather,'' in \textit{IEEE Sensors Journal}, vol. 24, no. 7, pp. 10515-10527, 1 April1, 2024.

\bibitem{ref32}
H. Hong, E. Pavez, A. Ortega, R. Watanabe and K. Nonaka, ``Motion Estimation And Filtered Prediction For Dynamic Point Cloud Attribute Compression,'' in \textit{2022 Picture Coding Symposium}, San Jose, CA, USA, 2022, pp. 139-143.

\bibitem{ref33}
M. Ghazal, A. Amer and A. Ghrayeb, ``Structure-Oriented Multidirectional Wiener Filter for Denoising of Image and Video Signals,'' in \textit{IEEE Transactions on Circuits and Systems for Video Technology}, vol. 18, no. 12, pp. 1797-1802, Dec. 2008. 

\bibitem{ref34}
K. Dabov, A. Foi, V. Katkovnik and K. Egiazarian, ``Image Denoising by Sparse 3-D Transform-Domain Collaborative Filtering,'' in \textit{IEEE Transactions on Image Processing}, vol. 16, no. 8, pp. 2080-2095, Aug. 2007. 

\bibitem{ref35}
H. Yuan, J. Liu, H. Xu, Z. Li and W. Liu, ``Coding Distortion Elimination of Virtual View Synthesis for 3D Video System: Theoretical Analyses and Implementation,'' in \textit{IEEE Transactions on Broadcasting}, vol. 58, no. 4, pp. 558-568, Dec. 2012.

\bibitem{ref36}
L. Wei, S. Wan, Z. Sun, X. Ding and W. Zhang, ``Weighted Attribute Prediction Based on Morton Code for Point Cloud Compression,'' in \textit{2020 IEEE International Conference on Multimedia \& Expo Workshops}, London, UK, 2020, pp. 1-6.

\bibitem{ref37}
C. Henry, L. Song and Z. Li, ``Fast Video Deduplication and Localization With Temporal Consistence Re-Ranking,'' in \textit{IEEE Transactions on Circuits and Systems for Video Technology}, vol. 34, no. 11, pp. 12006-12018, Nov. 2024.

\bibitem{ref38}
A. E. Jacob, N. Ashodariya and A. Dhongade, ``Hybrid search algorithm: Combined linear and binary search algorithm,'' in \textit{2017 International Conference on Energy, Communication, Data Analytics and Soft Computing}, Chennai, India, 2017, pp. 1543-1547.

\bibitem{ref39}
X. Zhang and W. Gao, ``Adaptive Geometry Partition for Point Cloud Compression,'' in \textit{IEEE Transactions on Circuits and Systems for Video Technology}, vol. 31, no. 12, pp. 4561-4574, Dec. 2021.

\bibitem{ref40}
G. J. Sullivan and T. Wiegand, ``Rate-distortion optimization for video compression,'' in \textit{IEEE Signal Processing Magazine}, vol. 15, no. 6, pp. 74-90, Nov. 1998.

\bibitem{ref41}
T. Wiegand, H. Schwarz, A. Joch, F. Kossentini and G. J. Sullivan, ``Rate-constrained coder control and comparison of video coding standards,'' in \textit{IEEE Transactions on Circuits and Systems for Video Technology}, vol. 13, no. 7, pp. 688-703, July 2003. 

\bibitem{ref42}
MPEG 3D Graphics Coding and Haptics Coding, ``Common Test Conditions for G-PCC,'' in \textit{ISO/IEC JTC1/SC29/WG07 MPEG output N00944}, Sapporo, July 2024.

\bibitem{ref43}
MPEG 3D Graphics Coding and Haptics Coding, ``Test model for geometry-based solid point cloud - GeS TM 7.0,'' in \textit{ISO/IEC JTC1/SC29/WG07 MPEG output N00965}, Sapporo, July 2024.

\bibitem{ref44}
Bjontegaard, G. . ``Calculation of average PSNR differences between RD-Curves.'' in \textit{ITU SG16 Doc. VCEG-M33}, 2001.

\end{thebibliography}
\end{document}